\documentclass[runningheads,a4paper]{llncs}
\pdfoutput=1

\usepackage{amssymb}
\usepackage{amsmath}
\usepackage{graphicx}

\graphicspath{{./figs/}}

\usepackage{booktabs}
\usepackage{subfigure}
\usepackage{url}
\urldef{\mailsa}\path|{alfred.hofmann, ursula.barth, ingrid.haas, frank.holzwarth,|
\urldef{\mailsb}\path|anna.kramer, leonie.kunz, christine.reiss, nicole.sator,|
\urldef{\mailsc}\path|erika.siebert-cole, peter.strasser, lncs}@springer.com|    
\newcommand{\keywords}[1]{\par\addvspace\baselineskip
\noindent\keywordname\enspace\ignorespaces#1}

\begin{document}

\mainmatter  

\title{On the Performance of Maximum Likelihood Inverse Reinforcement Learning}

\titlerunning{On the Performance of Maximum Likelihood Inverse Reinforcement Learning}

%
%
\author{H\'{e}ctor Ratia\inst{1} \and Luis Montesano\inst{1} \and Ruben Martinez-Cantin\inst{2}}

\authorrunning{H\'{e}ctor Ratia\and Luis Montesano\and Ruben Martinez-Cantin}

\institute{Universidad de Zaragoza
\and
Centro Universitario de la Defensa, Zaragoza
\path|hectorratia@gmail.com,{montesano,rmcantin}@unizar.es|
}

%
%

\toctitle{Lecture Notes in Computer Science}
\tocauthor{Authors' Instructions}
\maketitle

\begin{abstract}

  Inverse reinforcement learning (IRL) addresses the problem of
  recovering a task description given a demonstration of the optimal
  policy used to solve such a task. The optimal policy is usually
  provided by an expert or teacher, making IRL specially suitable for
  the problem of apprenticeship learning. The task description is
  encoded in the form of a reward function of a Markov decision
  process (MDP). Several algorithms have been proposed to find the
  reward function corresponding to a set of demonstrations. One of the
  algorithms that has provided best results in different applications
  is a gradient method to optimize a policy squared error
  criterion. On a parallel line of research, other authors have
  presented recently a gradient approximation of the maximum
  likelihood estimate of the reward signal. In general, both
  approaches approximate the gradient estimate and the criteria at
  different stages to make the algorithm tractable and efficient. In
  this work, we provide a detailed description of the different
  methods to highlight differences in terms of reward estimation,
  policy similarity and computational costs. We also provide
  experimental results to evaluate the differences in performance of the methods.

  \end{abstract}

\newcommand{\thetav}{\mathbf{\theta}}

\section{Introduction}
As proposed originally by Ng and Russell \cite{Ng00ICML}, the objective of the inverse reinforcement learning (IRL) problem is to determine the reward function that an \emph{expert} agent is optimizing in order to solve a task based on observations of that expert's behavior while solving the task. The motivation of IRL is twofold. First, it can provide computational models for human and animal behavior. It has been demonstrated that human action understanding can be modeled as inverse planning in a Markov decision process (MDP) model \cite{Baker2009,Ullman2010}. Second, it can be used for the design of intelligent agents, where the description of the tasks might not be easy to obtain. In the later, it is sometimes simpler to get demonstrations from an agent that already knows how to do the tasks \cite{Ziebart08aaai,Abbeel04icml,Neu07uai,Silva06icra}. For example, sometimes we are unable to describe a complex body movement, but we can show it so that the \emph{learner} can infer how to do it by herself \cite{fmelo07unified}. Furthermore, as discovered in \cite{Ratliff06icml} and exploited in \cite{Neu09ML}, there is a strong connection between the problem of IRL and the one of structured learning \cite{BakIr2007}.

Although the general IRL problem does not assume any model of the environment, the formal statement of the problem that appears in the literature assumes a MDP \cite{Puterman94} and that the \emph{expert} follows the principle of \emph{rational action}, that is, the expert agent always tries to maximize the reward function \cite{Baker2009}. This formulation is a generalization of the classical inverse optimal control (IOC)
problem in continuous domains \cite{Boyd1994,Krishnamurthy2010}. However, it is well known that, even with those restrictions, the problem of inverse reinforcement learning is ill-posed. That is, there is a virtually infinite number of rewards that accept the same demonstration as the optimal policy \cite{Ng99rewshap}. 

Thus, there are two ways of tackle the problem of IRL. One one hand, in the seminal paper of Ng and Russell \cite{Ng00ICML} and posterior works based on that one, such as \cite{Ramachandran07ijcai,Melo2010}, the authors try to characterize the space of solutions of the reward function. On the other hand, most of the recent algorithms care about the problem of \emph{apprenticeship learning}, where the learning is less concerned about the actual reward function, and the objective is to recover a policy that is \emph{close} to the demonstrated behavior \cite{Abbeel04icml,Neu07uai,Ratliff06icml,Ziebart08aaai,Syed08icml,Syed2007}. In that way, apprenticeship learning is related to imitation learning or learning by demonstration. Therefore, IRL in the case of apprenticeship learning includes a new restriction based on the \emph{similarity} or \emph{dissimilarity} of the expert and learner behaviors \cite{Neu09ML}. However, contrary to other approachers in imitation learning it can provide a more compact representation in terms of the reward function, it can generalize to states that have not been demonstrated and it is more robust to changes in the environment. 

This papers studies an algorithm for maximum likelihood IRL from two different points of view. First, it shows its connections to prior work on convex optimization for IRL, namely with the use of gradient methods in a least-squares optimization. Second, it provides a comparison of the performance of maximum likelihood against other IRL methods as well as of an approximation of the gradient. The results show that maximum likelihood always obtains, for the studied problems, the best results. Moreover, the solutions with the gradient approximation do practically not degrade while achieving a considerable speed up in computational time.	

The reminder of the paper is organized as follows. After introducing the notation and inverse reinforcement learning methods in Section \ref{sec:prelim}, Section \ref{sec:GIRL}  describes maximum likelihood IRL and discusses its connections to other algorithms. Section \ref{sec:results} presents the experimental results. Finally, in Section \ref{sec:conclusions} we draw the conclusions.

\section{Preliminaries}
\label{sec:prelim}
In this section we will introduce the notation used in this article, point to the basic equations we need from direct reinforcement learning, and enunciate the inverse reinforcement learning problem.

\subsection{Markov decision processes}
A Markov decision process (MDP) is a tuple $(\mathcal{X}, \mathcal{A}, P, \gamma, R)$ where
\begin{itemize}
\item $\mathcal{X}$ is a set of states, 
\item $\mathcal{A}$ is a set of actions, 
\item $P(x'\mid x, a) \equiv P_{x'ax}$ is the probability of transitioning to state $x' \in \mathcal{X}$ when taking action $a \in \mathcal{A}$ in state $x \in \mathcal{X}$, i.e., $P: \mathcal{X} \times \mathcal{A} \times \mathcal{X} \rightarrow [0,1]$,
\item $R$ is a reward function. $R(x,a) \equiv R_{xa}$ returns the reward for taking action $a$ in state $x$. $R: \mathcal{X} \times \mathcal{A} \rightarrow \mathbb{R}$ and
\item $\gamma \in [0,1)$ is the discount factor.
\end{itemize}

The purpose of the MDP is to find the action sequence that maximizes the expected future reward:
\begin{equation}
	V(x)=\mathbb{E} \left [ \sum_{t=0}^{\infty} \gamma^t R(x_t,a_t) \middle \vert  x_0=x \right ]
\label{eq:valuefunction}
\end{equation}

The sequence of actions is encoded in a \emph{policy}, which is a mapping $\pi :\mathcal{X} \times \mathcal{A} \rightarrow [0,1]$ such that $\pi(x,a)=P(a\mid x)$. In the case of a deterministic policy, the probability distribution collapses to a single action value.
We can associate a \emph{value function} with a particular policy $\pi$, $V^{\pi}(x)=\mathbb{E}_{\pi} [ \sum_{t=0}^{\infty} \gamma^t R(x_t,a_t) \vert x_0=x ]$, where the expectation also considers the stochasticity in the policy. Then, the optimal policy $\pi^*$ is defined as the policy such that the associated value function $V^{*}(x)$ is greater or equal than the value function of any other policy, that is $V^{*}(x) = \sup_\pi V^{\pi}(x)$. The optimal value function satisfies the Bellman equations:
\begin{equation}
\label{eq:bellmanv}
	V^*(x)=\max_{a\in\mathcal{A}} \left [ R(x,a) + \gamma \sum_{y\in\mathcal{X}} P(y\mid x,a) V^*(y) \right ]
\end{equation}  
We can also associate an action-value function, or \emph{Q-function}, with each policy
\begin{equation}
	Q^{\pi}(x,a)=\mathbb{E}_{\pi} \left [ \sum_{t=0}^{\infty} \gamma^t R(x_t,a_t) \middle \vert x_0=x, a_0=a \right ]
\end{equation}
where $a_t$ is generated by following policy $\pi$ for $t>0$. The Q-function can also be updated from the Bellman equation:
\begin{equation}
\label{eq:bellmanq}
	Q^*(x,a)= \left [ R(x,a) + \gamma \sum_{y\in\mathcal{X}} P(y\mid x,a) V^*(y) \right ]
\end{equation}
Therefore, the optimal policy can be computed as
\begin{equation}
	\pi^*(x)= \arg \max_{a\in\mathcal{A}} Q^*(x,a)
\label{eq:greedypi}
\end{equation}

\subsection{Differentiable Markov decision processes}

One of the limitations for the Bellman formulation of MDPs is the non-differentiability of the maximum function. Thus, in many algorithms replace the maximum function by a softmax function \cite{Ziebart__2010_6590}. For example, we can replace equation \eqref{eq:greedypi} by the softmax version of it, the Boltzmann policy:
\begin{equation}
  \label{eq:boltpi}
  	\pi^*(x)=\frac{e^{\frac{Q^*(x,a)}{\eta}}}{\sum_{a}e^{\frac{Q(x,a)}{\eta}}}
\end{equation}
where $\eta$ is the Boltzmann temperature. If the temperature $\eta \rightarrow 0^+$, then equation \eqref{eq:boltpi} becomes equivalent to equation \eqref{eq:greedypi}. If $\eta \rightarrow \infty$ then $\pi(x)$ becomes a uniform random walk. 

\subsection{Inverse reinforcement learning}
\label{sec:irl}
As stated before, the inverse reinforcement learning (IRL) problem
consists of learning the reward function of a reward-less MDP, that is, MDP\textbackslash $R$, given a set of trajectories from an expert or teacher. Formally, in the IRL problem we are given a dataset 
\begin{equation}
 \mathcal{D} =\left\{ \left( x_i , a_i \right)\right\}_{i=1}^M  
\end{equation}
containing observations of an expert agent acting in the MDP
$\mathcal{M} = ( \mathcal{X}, \mathcal{A}, P, R, \gamma )$. The dataset can contain full or partial trajectories of the expert, or even some sparse action selections.

We assume the expert acts near-optimally trying to solve a certain task encoded by the reward function $R$. Both the task and the reward signal are unknown to the learning agent. Her goal is to find a reward function that explains the observed behavior of the expert.

The problem of IRL is, by definition, ill-posed \cite{Ng00ICML} because different rewards can produce the same behavior \cite{Ng99rewshap}, that is, a demonstration cannot generate a single reward signal, neither discriminate among an infinite set of reward functions. Furthermore, the set of solutions for a demonstration include degenerate cases such a flat reward for every action-state pair \cite{Ng00ICML}. Also, there can be some rewards that do not depend directly on the state of the system, but on some intrinsic parameters of the agent \cite{Singh04intrRL}; or the agent might not fully observe the system, in which case the direct problem becomes a partially observable MDP (POMDP) \cite{Cassandra1994,Chot2011}.

At this point is important to note that, apart from the sums in the value function, there is no other assumption in this paper about whether the spaces are discrete or continuous. In fact, many algorithms that were designed for discrete spaces have been applied in continuous setups, provided that there is a planning algorithm to solve the direct problem just by replacing the sums by the corresponding integrals \cite{Abbeel2008,Krishnamurthy2010,Ziebart__2010_6590}.

\subsubsection{Inverse reinforcement learning as convex optimization}

As presented in \cite{Neu09ML}, many algorithms for apprenticeship learning based on IRL, share a common formulation\cite{Abbeel04icml,Neu07uai,Ratliff06icml,Syed2007,Boularias2010}. Basically, the objective is to find the reward that minimizes the similarity between the expert's and learner's behavior:
\begin{equation}
  \label{eq:dissimilarity}
  R^* = \arg \max_R J(\pi_R,\mathcal{D})
\end{equation}
Those algorithms also include some extra assumptions to constrain the admissible reward set in a way to reduce the effects of being an ill-posed estimation. The most extended assumption is to consider that the reward function is a linear combination of basis functions $\phi(x,a)$, also called, state features. 
\begin{equation}
  \label{eq:features}
  R_{\theta}(x,a) = \sum_{i=1}^N\theta_i \phi_i(x,a)
\end{equation}
where $\theta$ is a vector of feature weights. Then equation \eqref{eq:valuefunction}, can be rewritten in matrix form as $V(x)=\mathbf{\theta} \ \Phi^\pi(x,a)$, where
\begin{equation}
  \label{eq:expfeat}
  \Phi^\pi_i(x,a) = \mathbb{E}_\pi\left[\sum_{t=0}^\infty \gamma^t \phi_i(x,a) \middle \vert x_0=x \right]
\end{equation}
The selection of the features might be tricky depending on the problem. In some works, the IRL problem is formulated by directly trying to find an arbitrary $R(x,a)$. This is equivalent to use the indicator function as feature $\phi_i(x,a)=\mathbb{I}_i(x,a)$. Some authors also have tried to learn the features as a part of the IRL problem\cite{Krishnamurthy2010,Levine2010}.

As pointed out by \cite{Ratliff06icml} this estimation resembles the problem of structured learning \cite{BakIr2007}. The main hypothesis of \cite{Taskar2003} is that, for a certain kind of combinatorial problems in the form of equation \eqref{eq:features}, computing the likelihood function is intractable. Therefore, they propose the max-margin method, which approximates the correct solution with polynomial complexity. In fact, the work of \cite{Neu09ML} shows that under certain conditions, the problems of structured learning and IRL are equivalent.
In this paper, we show that for IRL problems we can obtain an alternative good approximation of the likelihood function with better performance than the heuristic proposed for structured learning.

\subsubsection{Inverse reinforcement learning as probabilistic inference}

The problem of IRL can also be solved using Bayesian inference. Ramachandran and Amir \cite{Ramachandran07ijcai} presented an algorithm called Bayesian inverse reinforcement learning, where they consider that the unknown reward function as an stochastic variable which can be inferred based on the observations from the demonstration $  P(R \mid \mathcal{D})  \propto P(\mathcal{D} \mid R) P(R) $. Then, they introduce the following likelihood model
\begin{equation}
  \label{eq:rewardlik}
  P(\mathcal{D} \mid R) = \prod_i P(x_i,a_i \mid R) \propto e^{\alpha \sum_i Q^*(x_i,a_i)}
\end{equation}
where $\alpha$ is a parameter of the distribution that represents the confidence on the expert. The likelihood of each pair $(x,a)$ is equivalent to the Boltzmann policy from equation \eqref{eq:boltpi} assuming that the confidence is the inverse of the Boltzmann temperature $\alpha = 1/\eta$. Although for apprenticeship learning, the computation of a full distribution of rewards might seem excessive, it has some advantages as being able to use the uncertainty in the estimation in an active learning framework \cite{macl09airl}.

\subsubsection{Inverse reinforcement learning as density estimation}

Although they are not within the scope of the paper, it is worth mentioning a set of algorithms rooted on the computation of the KL-divergence (or relative entropy) between the optimal policy and the passive dynamics of the system \cite{Boularias2011,Krishnamurthy2010,Ziebart08aaai}. In contrast with the two approaches described above, this family of algorithms does not need to solve the direct planning problem. Instead, they require to compute the distributions of trajectories following the passive dynamics of the system and the potentially optimal dynamics of the expert. Furthermore, it is not clear if those algorithms are comparable to other IRL algorithms since the problems they solve are different. For example, \cite{Neu09ML} shows that the dissimilarity function that those algorithms are optimizing do not use optimal policies. Besides, \cite{Krishnamurthy2010} shows that those methods are within the framework of linearly-solvable MDPs.


\section{Maximum likelihood IRL}
\label{sec:GIRL}

In this section we describe in detail the algorithm introduced in \cite{macl09airl} to solve the IRL problem using an estimate of the gradient of the likelihood function from equation \eqref{eq:rewardlik}. Then, we present new connections between that algorithm and other algorithms in the literature. First, to provide a uniform framework, we will use the reward parametrization of equation \eqref{eq:features}. As commented before, the original formulation of \cite{macl09airl} can be recovered by taking $\phi_i(x,a)=\mathbb{I}_i(x,a)$.

For simplicity, we assume that the feature functions are known in advance. Therefore, the reward function is fully determined by the vector of weights $\theta$. We define the likelihood of the data set as the product of the likelihood of the state-action pairs $ P(x_i,a_i \mid \mathcal{D}) = \ell_\theta(x_i,a_i) $ defined as in equation \eqref{eq:rewardlik}, 
\begin{equation}
   \label{eq:likelihood}
   \mathcal{L}_\theta (\mathcal{D})= \prod_{i=1}^{M}\ell_\theta(x_i,a_i)  
\end{equation}
where $M$ is the number of demonstrated pairs.

Thus, a gradient ascent algorithm can be used to estimate the reward function $R$ maximizing the log-likelihood function with respect to the demonstrations $\mathcal{D}$. As described above, we are interested in computing a reward function $R^*$ such that:
\begin{equation}
  \label{eq:maxlogl}
  R_{\theta}^*= \arg\max_R \log \mathcal{L}_\theta (\mathcal{D})  
\end{equation}
subject to the constraints on the parameter weights $\theta_i \geq 0$ and $\|{\theta}\|_1=1$, where
\begin{equation}
	\log \mathcal{L}_\theta (\mathcal{D})= \sum_{i=1}^{M} \log ( \ell_\theta(x_i,a_i))
	\label{eq:logl}
\end{equation}

The model in Eq. \ref{eq:rewardlik} assumes that the probability of the expert choosing an action is proportional to the action's value; e.g. actions with higher $Q^*$ are more likely to be selected. Given an observed pair $(x,a)$, the likelihood of the pair under the reward function $R_{\theta}$ is defined as in equation \eqref{eq:rewardlik}, which is equivalent to the Boltzmann policy from equation \eqref{eq:boltpi}

\subsection{Computing the gradient}
At this point, we need to determine the gradient
\begin{equation}
  \label{eq:updaterule}
\mathbf{\nabla}_\thetav \log \mathcal{L}_\thetav (\mathcal{D}) = \frac{\partial}{\partial \thetav}\left [ \sum_{i=1}^{M} \log( \ell_\thetav(x_i,a_i)) \right ] = \sum_{i=1}^{M} \frac{1}{\ell_\thetav(x_i,a_i)}\frac{\partial \ell_\thetav}{\partial \thetav}(x_i,a_i)
\end{equation}

The partial derivative of the pair likelihood can be calculated as:
\begin{equation}
  \label{eq:likpair}
\frac{\partial \ell_\theta}{\partial \theta_k}(x,a)=\frac{\ell_\theta(x,a)}{\eta} \left(\frac{\partial Q^*}{\partial \theta_k}(x,a)-\sum_{b \in \mathcal{A}} \ell_\theta(x,b)\frac{\partial Q^*}{\partial \theta_k}(x,b)\right)  
\end{equation}
The likelihood of the pair $(x_i,a_i)$ can be easily calculated by solving the direct RL problem. However, we still need to find the gradient of the optimal \emph{Q-function} with respect to the reward parameters $\theta$. This derivative is not trivial, since there is double dependency of $R_\theta$ in $Q^*$, first through the accumulated reward and second through the optimal policy. However, we can find an estimate of the derivatives of the $Q^*$ functions. 


\subsubsection{Fixed-point estimate}
As shown by \cite{Neu07uai} the derivatives of the $Q^*$ functions can be computed \emph{almost everywhere} with a fixed point equation:
\begin{equation}
  \label{eq:fixedpoint}
  \psi_\theta(x,a) = R'_\theta(x,a) + \gamma \sum_{y\in\mathcal{X}} P(y\mid x,a) \sum_{b\in\mathcal{A}} \pi(b,y) \psi_\theta(b,y)
\end{equation}
where $R'_\theta(x,a)$ is the derivative of the reward w.r.t. $\theta$ and $\pi$ is any policy that is greedy with respect to $Q_\theta$. 
If the reward is in the linear form of \eqref{eq:features}, then $R'_\theta(x,a) = \phi(x,a)$ and the solution to that equation is also the set of feature expectations $\Phi_\pi(x,a)$ defined in equation \eqref{eq:expfeat}.
Thus, the resulting equation for the derivative of $\ell_\theta$ is
\begin{equation}
\frac{\partial \ell_\thetav}{\partial \theta_k}(x,a)=\frac{\ell_\thetav(x,a)}{\eta} \left(\Phi^\pi_k(x,a)-\sum_{b \in \mathcal{A}} \ell_\thetav(x,b)\Phi^\pi_k(x,b)\right)	\label{partialderivative}
\end{equation}

\subsubsection{Independence assumption}
The second estimate is an approximation based on the assumption that the \emph{policy remains unchanged under a small variation in the reward function} as in \cite{macl09airl}. The Bellman recursion of the value-function and Q-function from equations \eqref{eq:bellmanv}  and \eqref{eq:bellmanq} using vector notation are:
\begin{align*}
V^*&= \max_{a \in \mathcal{A}} \left[ R_a+\gamma P_a V^* \right] \;\; &V^\pi= R_\pi+\gamma P_\pi V^\pi \\
Q^*_a&=R_a+\gamma P_a V^* \;\; &Q^\pi_a = R_a + \gamma P_a V^\pi
\end{align*}
then, we notice that for any optimal policy $\pi^*$:
\begin{equation}
V^*=\left ( \mathbf{I} -\gamma P_{\pi^*} \right )^{-1} R_{\pi^*}
\end{equation}
These expressions can be combined into
\begin{equation}
Q^*_a=R_a+\gamma P_a \left ( \mathbf I -\gamma P_{\pi^*} \right )^{-1} R_{\pi^*} 
\end{equation}
Let us define $\mathbf{T}=\mathbf{I} -\gamma P_{\pi^*}$. Ignoring the dependency on the policy of the right hand side of the equation one obtains the following approximation of the derivative
\begin{equation}
\frac{\partial Q^*_a}{\partial \theta_k}=\phi_k (x,a)+\gamma P_a \mathbf T^{-1} \left [ \sum_{b \in \mathcal{A}} \pi^*(x,b) \phi_k(x,b)  \right ]_x \; .
\label{eq:appderiv}
\end{equation}
In contrast with the fixed point method of \cite{Neu07uai}, this approximation has a computational cost that is polynomial (due to the inverse) in the number of states. In the experiments we show that both methods provide comparative results in terms of accuracy.

\subsection{Comparison to other IRL algorithms}
Although we have presented the maximum likelihood algorithm in terms of probabilistic inference, it can be easily reformulated in terms of convex optimization, in order to be comparable to other IRL algorithms as in \cite{Neu09ML}. For example, the update rule in \eqref{eq:updaterule} can be rewritten by replacing the sum over the dataset for a sum over all state-action pairs:
\begin{equation}
  \label{eq:newupdaterule}
\begin{split}
  \Delta_k &=\sum_{i=1}^{M} \frac{1}{\ell_\theta(x_i,a_i)}\frac{\partial \ell_\theta}{\partial \theta_k}(x_i,a_i)\\
&=\sum_{x,a \in \mathcal{X} \times A} M \; \mu_E(x) \; \hat{\pi}_E(a\mid x)\frac{1}{\ell_\theta(x,a)}\frac{\partial \ell_\theta}{\partial \theta_k}(x,a)
\end{split}\end{equation}
where $\mu_E(x)$ is the observed state visitation frequencies of the expert's behavior\footnote{Remember that the indicator function is defined as $\mathbb{I}_{xa}(y,b) = \left\{ \begin{array}{cc} 1 & \mbox{if} \; x=y \wedge a=b\\ 0  & \mbox{otherwise}\end{array}\right.$.} 
\begin{equation}
\mu_E(x)=\frac{\sum_{i=1}^{M} \mathbb{I}(x_i=x)}{M}
\end{equation}
and $\hat{\pi}_E(a\mid x)$ the policy estimated from observations of the expert's behavior
\begin{equation}
\hat{\pi}_E(a\mid x)=\frac{\sum_{i=1}^{M} \mathbb{I}(x_i=x \wedge a_i=a)}{\sum_{i=1}^{M} \mathbb{I}(x_i=x)}.
\end{equation}
As shown in \cite{Boularias2010}, the estimated expert policy can be inaccurate when there are many unvisited states or the demonstrations are scarce. When no observations are available for a state we will assume that the optimal policy is a random walk.

In terms of maximization, the constant $M$ can be dropped and we can replace the likelihood function for the Boltzmann policy since, by definition, $\ell_\theta(x,a)=\pi_\theta(a\mid x)$. Therefore, we obtain:
\begin{equation}
\Delta_k=\sum_{x,a \in \mathcal{X} \times \mathcal{A}} \mu_E(x) \hat{\pi}_E(a\mid x)\frac{1}{\pi_\theta(a\mid x)}\frac{\partial \pi_\theta}{\partial \theta_k}(a\mid x)
\label{eq:update2}
\end{equation}
which can be integrated to obtain the \emph{similarity} function being maximized:
\begin{equation}
  \label{eq:dissimilaritygirl}
  J(\pi_\theta,\mathcal{D}) =\sum_{x,a \in \mathcal{X} \times \mathcal{A}} \mu_E(x) \hat{\pi}_E(a\mid x) \log \pi_\theta(a\mid x)
\end{equation}

The maximum likelihood approach, therefore, describes an alternative cost function for IRL problems. 
Indeed, following \cite{Neu09ML}, it belongs to the the family of algorithms aiming to match the policies instead of the feature expectations as the policy matching algorithm described in \cite{Neu07uai}. However, the latter uses a least square cost function instead of a maximum likelihood approach.


\section{Experiments}
\label{sec:results}

In this section we evaluate the performance of the maximum likelihood IRL, which we called GIRL for Gradient-based-IRL, and compare it with other IRL algorithms: Policy Matching (PM) \cite{Neu07uai} and the Multiplicative Weights Algorithm (MWAL) \cite{Syed2007}. The experiments are divided in two different scenarios. First, we use a standard grid world as used in many IRL papers since Abbeel and Ng \cite{Abbeel04icml}. Then, we use the sailing simulator proposed by Vanderbei \cite{Vanderbei} and used for IRL in Neu and Szepesvary \cite{Neu07uai}. 

In addition to compare the methods, we also evaluate the impact of approximating the derivative using Eq. \ref{eq:appderiv}. Since for the reward model used in the paper the derivative is equal to the feature expectations, it is possible to compute them using the same approximation for the MWAL algorithm. In order to have a fair comparison in terms of computational time, we also report results where the feature expectations (i.e. the derivative) are estimated in a single step (horizon one). We will denote the full fixed point recursion as FP, the Independence assumption of Eq. \ref{eq:appderiv} as IA and the one step fixed point as FP1.

First we offer a description of the metrics that we are going to use to evaluate the algorithms among them. Then, we will describe the setup used for the experiments and show some results. Finally, we provide some quantitative results about the different criteria to approximate the derivative of the Q-function.

\subsection{Performance metrics}

We are interested in comparing the ability of the algorithms to a) recover the structure of the problem and b) propose policies that perform as good as the demonstrated expert. 

One measure of performance is the accumulated rewards using policy evaluation \cite{Sutton98}. The value-function following policy $\pi$ from a state $x$ with a reward function $R$ is
\begin{equation}
  V_R^\pi(x)=\mathbb{E}_{\pi}\left\{R_{t+1}+\gamma R_{t+2}+ \gamma^2 R_{t+3}+ \ldots \mid x_t=x\right\}
\end{equation}
The total value of a policy $\pi$ is:
\begin{equation}
V_R^\pi=\sum_{x \in \mathcal{X}} V_R^\pi(x) P(x_0=x)
\end{equation}
We call $R_E$ to the real reward which the expert is optimizing, and $\pi_E$ to the policy of the expert. The IRL algorithm converges to reward function $R_{IRL}$ with corresponding optimal policy  $\pi_{IRL}$. Then, the maximum obtainable accumulated reward is $V_{R_E}^{\pi_E}$ and the accumulated reward obtained by the IRL solution $V_{R_E}^{\pi_{IRL}}$.

Another measure of how good the algorithm solves the problem is the comparison of the greedy policies of the expert and the learner. We measure what fraction of states of the greedy version of the learned policy $\pi_{IRL}$ match the actual optimal greedy policy $\pi_E$. However, this measure can be sometimes misleading, since taking the wrong actions in critical parts of the problem can be disastrous reward-wise. Also in some parts of the problems there is more than one optimal action. This has been called the label bias \cite{Ziebart08aaai} problem and the desired performance, whether to map the optimal policy or the distribution over paths, depends on the task.

\begin{figure}
  \centering
\begin{tabular}{cccc}
\includegraphics[width=0.23\textwidth]{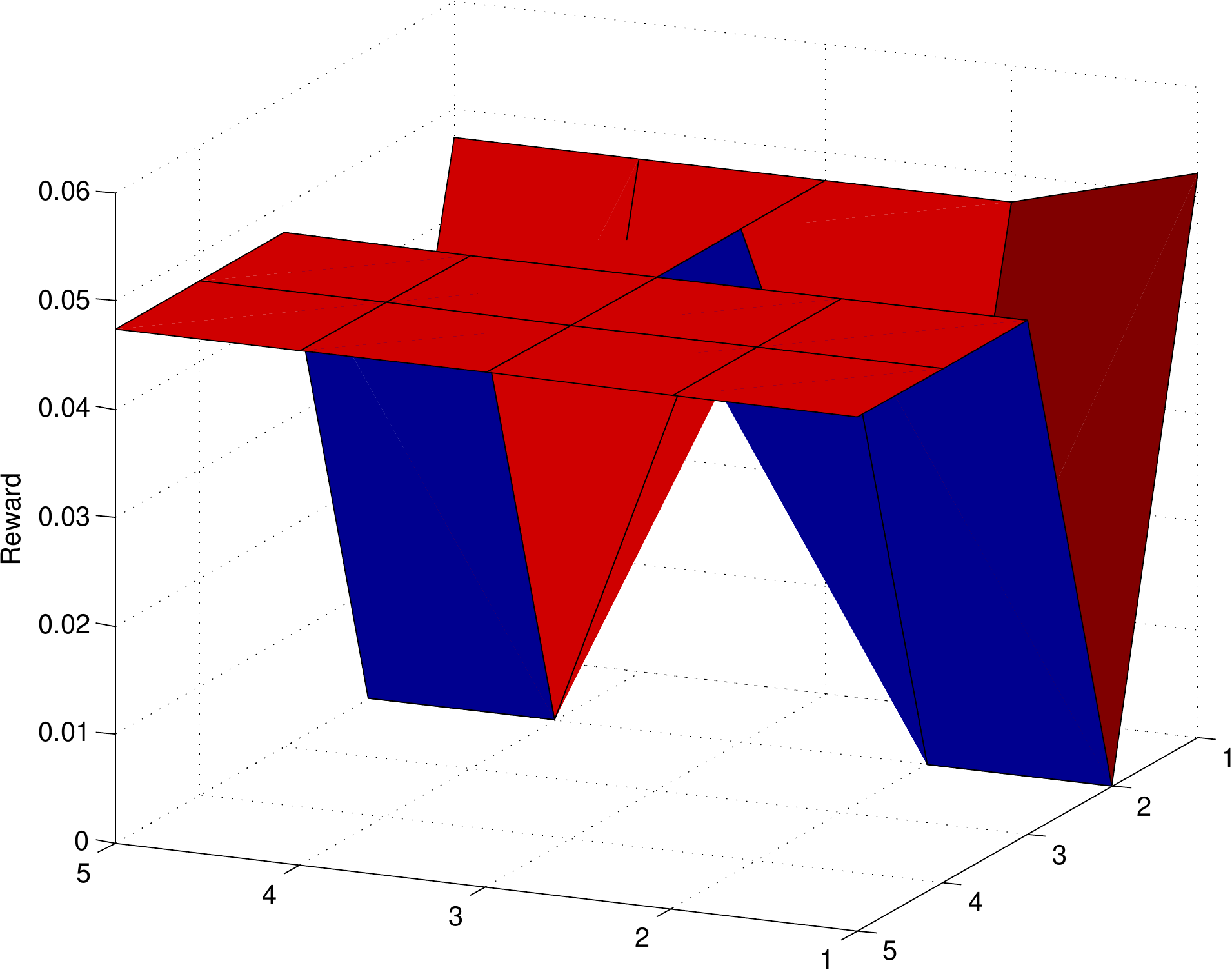} &
\includegraphics[width=0.23\textwidth]{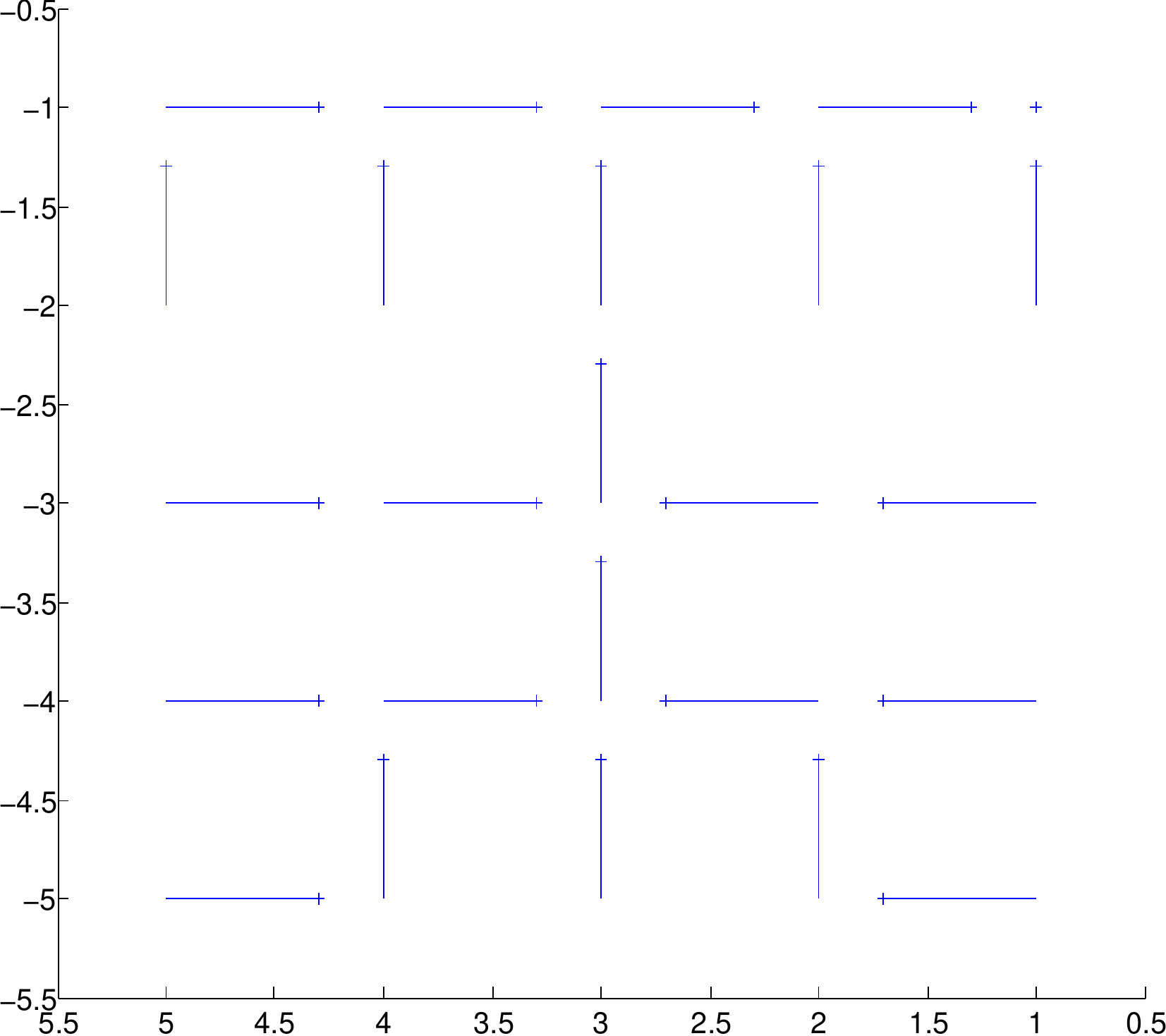} &
\includegraphics[width=0.23\textwidth]{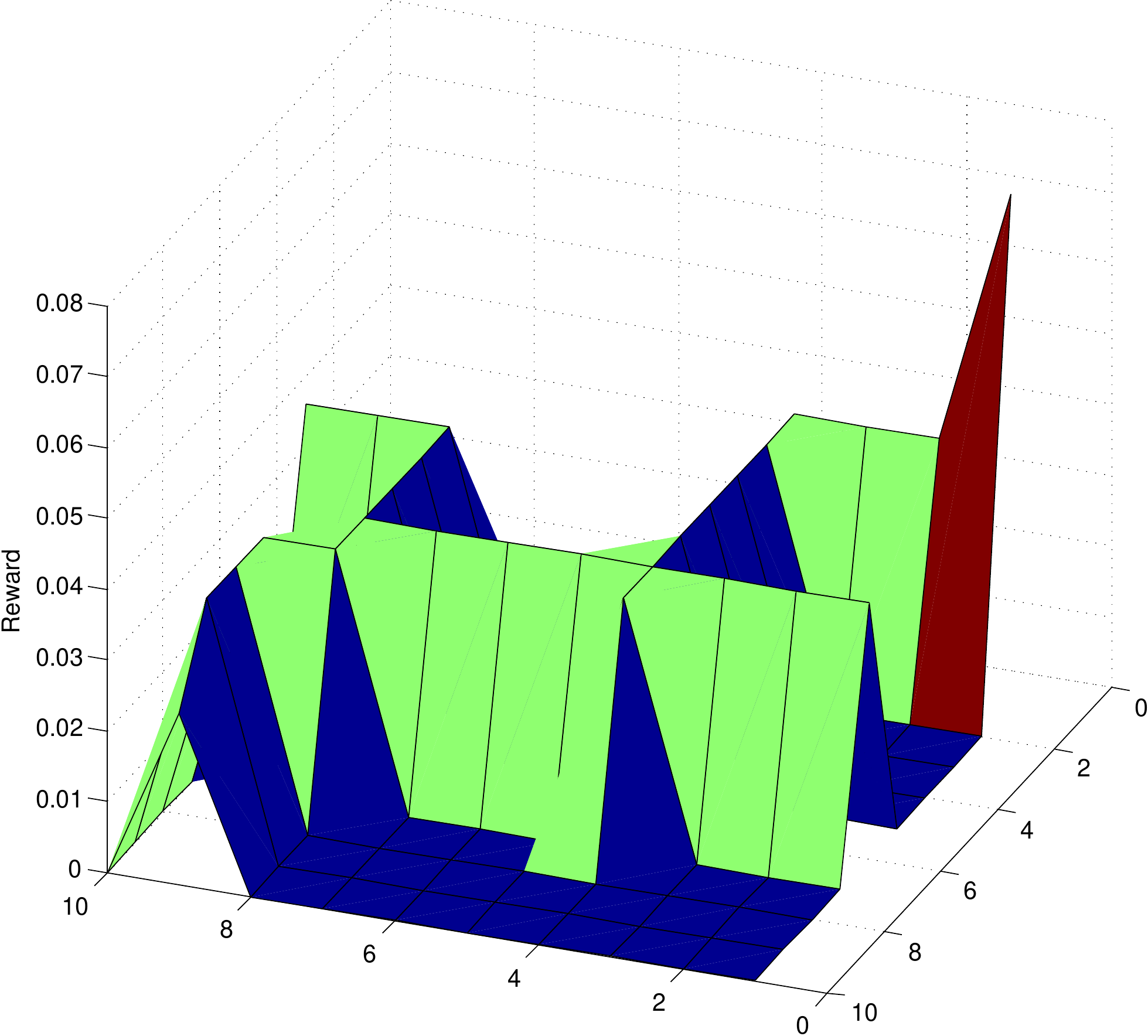} &
\includegraphics[width=0.23\textwidth]{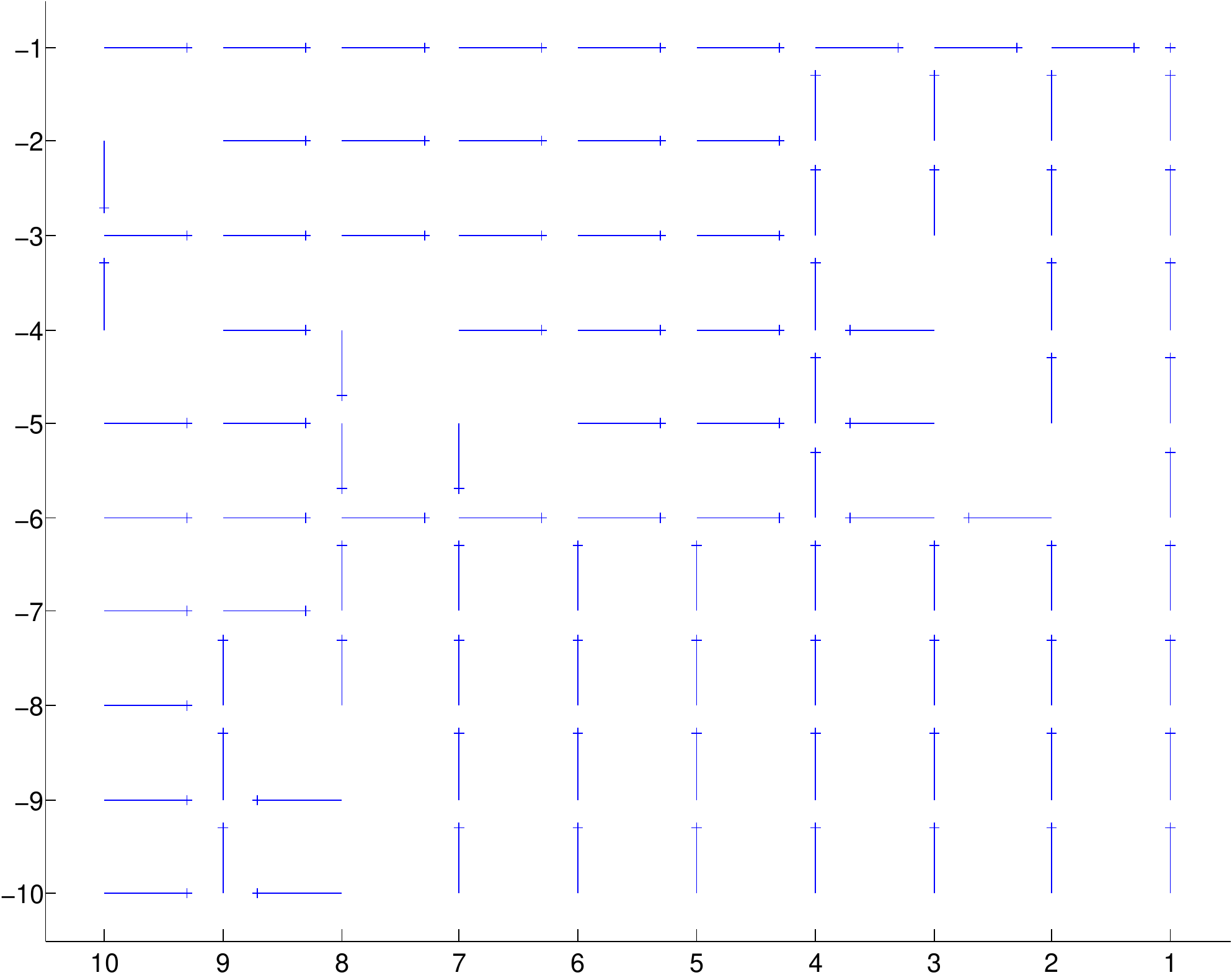}\\
(a) & (b) & (c) & (d) 
\end{tabular}
\caption{Narrow passage problem: (a) the true reward, (b) the expert  policy. Paths problem: (c) the true reward, (d) the expert  policy}  
\label{fig:problems}
\end{figure}

\subsection{Grid world}

The grid world is made up of grid squares with five actions available to the agent, corresponding to moves in the compass directions (N,S,E,W) plus an action for staying in the actual state. We assume that the real reward is in the linear form from equation \eqref{eq:features}. The grid is divided in macro-cells $\Psi_i$ which can span several grid squares. The reward features are the indicator function of the agent's state being inside a macro-cell, $\mathbb{I}_\Psi(x)$. Note that in this example, the real and estimated rewards depend only on the state $R(x,a) = R(x)$.

After doing some preliminary tests, we found that it is important for the grid world example to have some structure, in order to be able to draw conclusions about IRL algorithms. In many papers, the reward is chosen randomly resulting in problems with little structure. In those cases, the IRL algorithm needs only to find the goal point to perform well thus not helping to properly evaluate the algorithms. Based on this observation, we propose two different grid world problems with structure to asses performance:

\begin{itemize}
	\item A narrow passage problem, see Figures \ref{fig:problems} (a) and (b), where one corner has slightly better reward than the other states but to get to the corner the agent has to be careful not to fall in the pit and walk along the narrow passage. This problem has 25 macro-cells arranged in a 5x5 square grid. The true reward accumulated by the expert is $V_{R_E}^{\pi_E}=0.9964$.
	\item A path following problem, see Figures \ref{fig:problems} (c) and (d), where most states give no reward but the goal state and the states following some paths to this goal. The agent has to follow these paths and beware of stepping outside on the way to the goal. This problem has 100 macro-cells arranged in a 10x10 square grid. The true reward accumulated by the expert is $V_{R_E}^{\pi_E}=0.9358$
\end{itemize}

\begin{figure}[!t]
 \centering
\subfigure[Policy similarity, FP]{\includegraphics[width=0.3\textwidth]{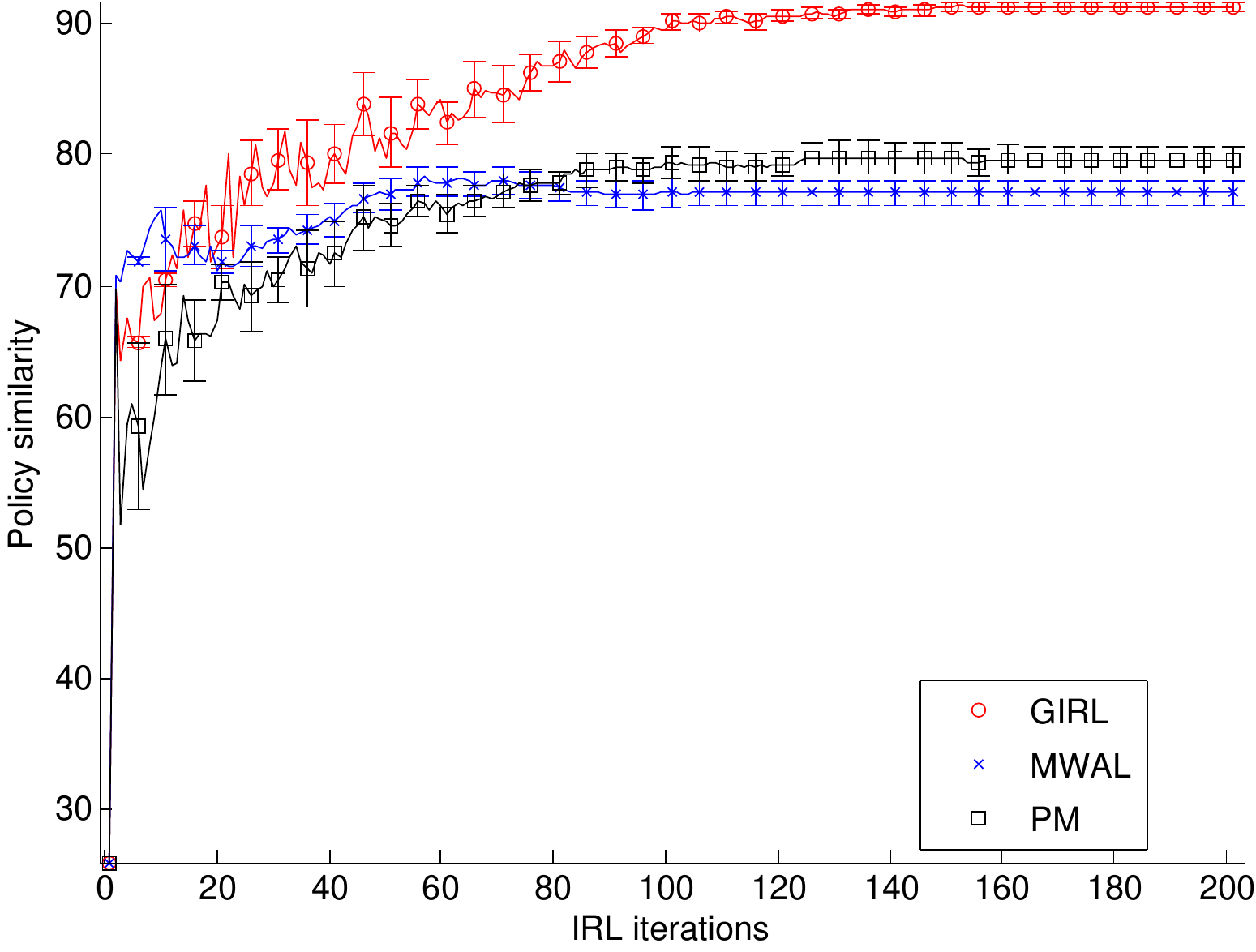}}
\subfigure[Policy similarity, IA]{\includegraphics[width=0.3\textwidth]{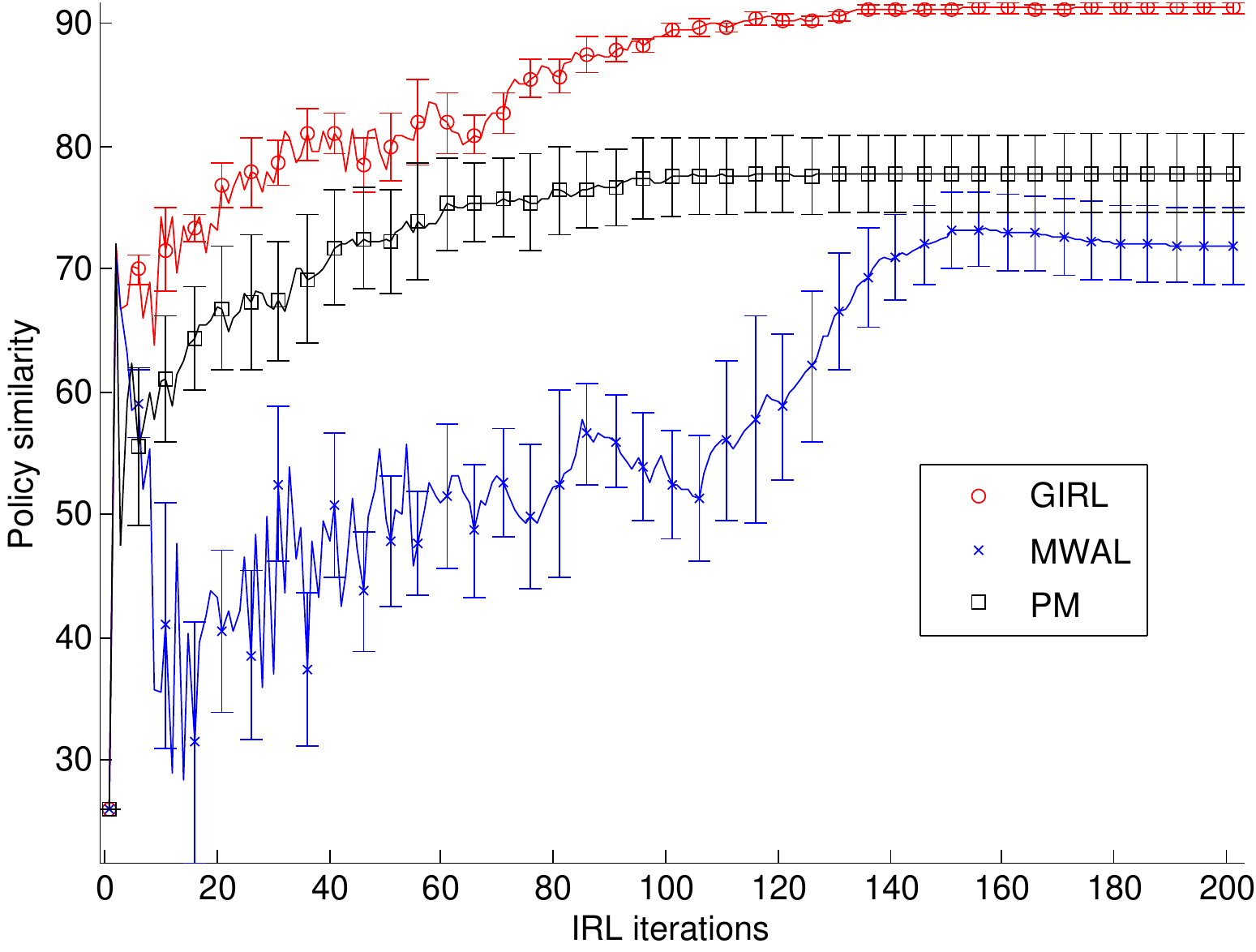}}
\subfigure[Policy similarity, FP1]{\includegraphics[width=0.3\textwidth]{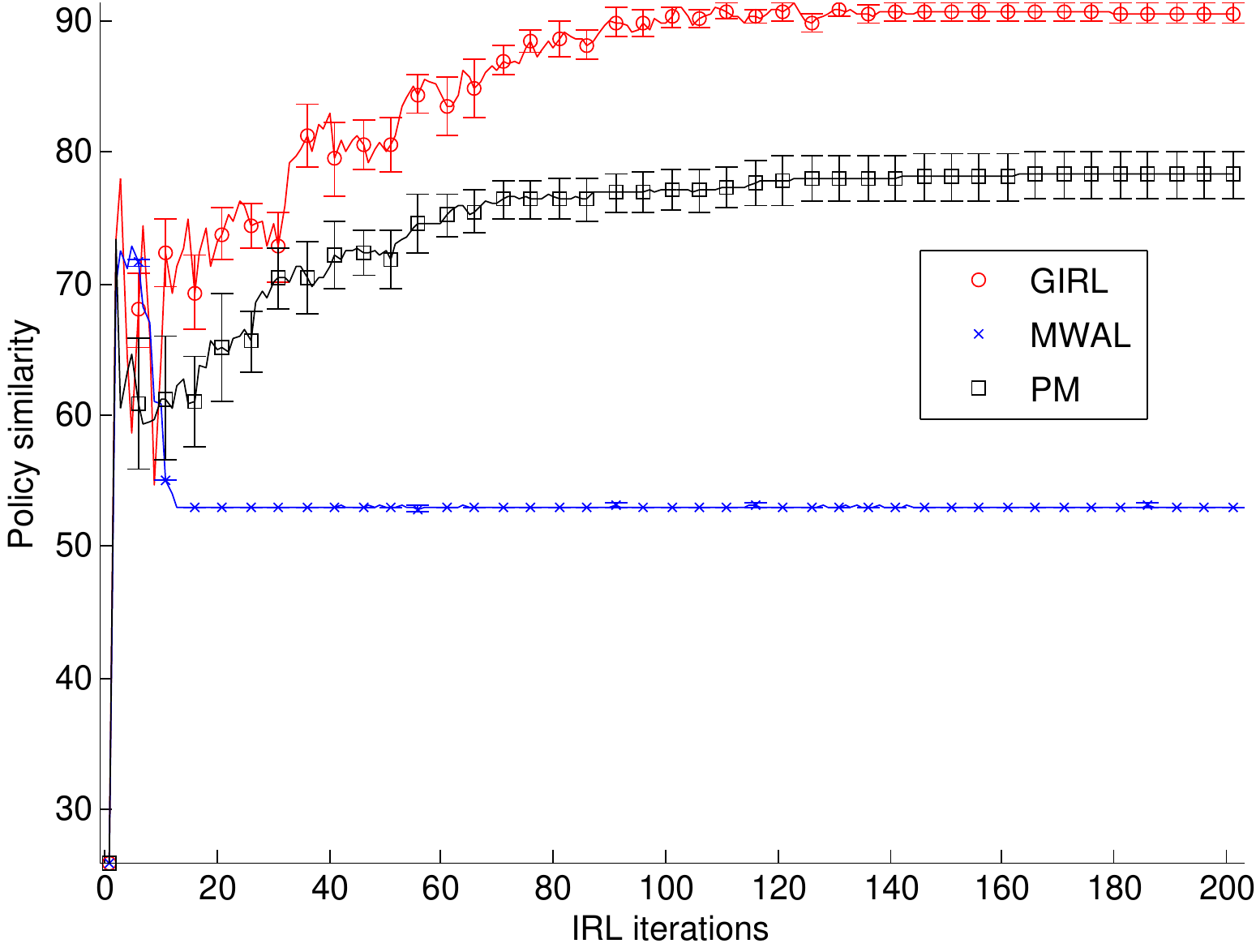}}

\subfigure[Acc. reward, FP]{\includegraphics[width=0.3\textwidth]{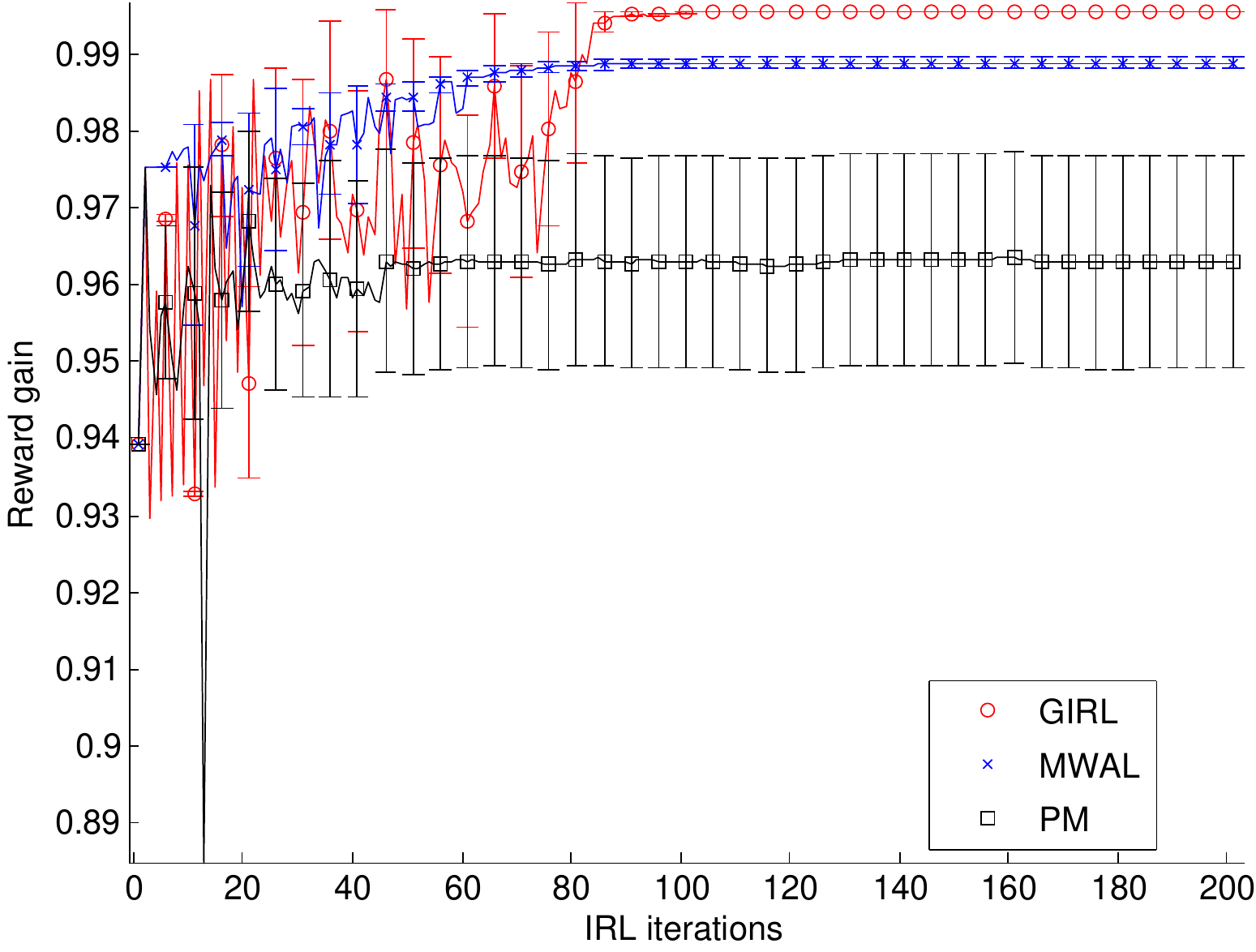}}
\subfigure[Acc. reward, FP]{\includegraphics[width=0.3\textwidth]{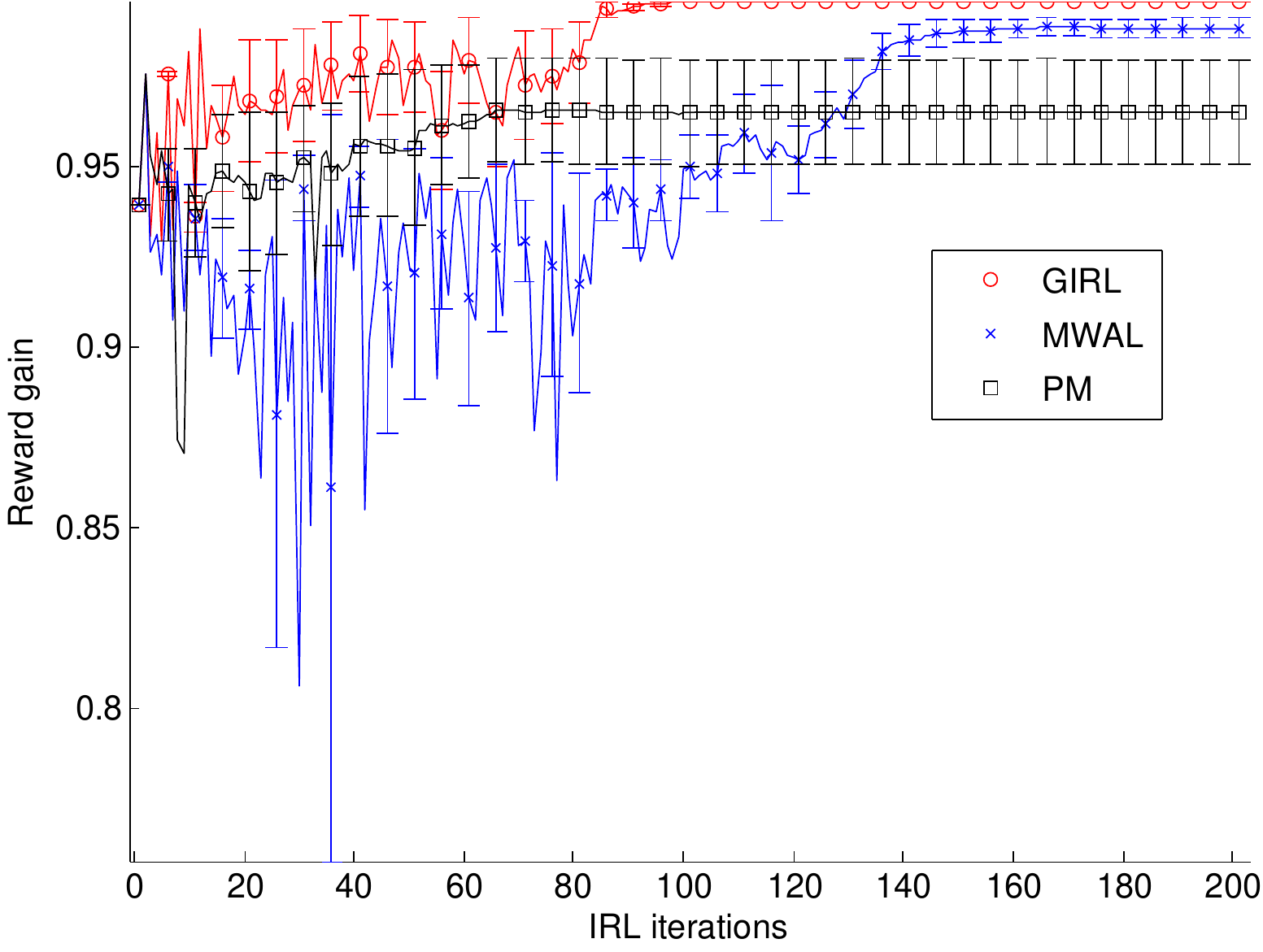}}
\subfigure[Acc. reward, FP1]{\includegraphics[width=0.3\textwidth]{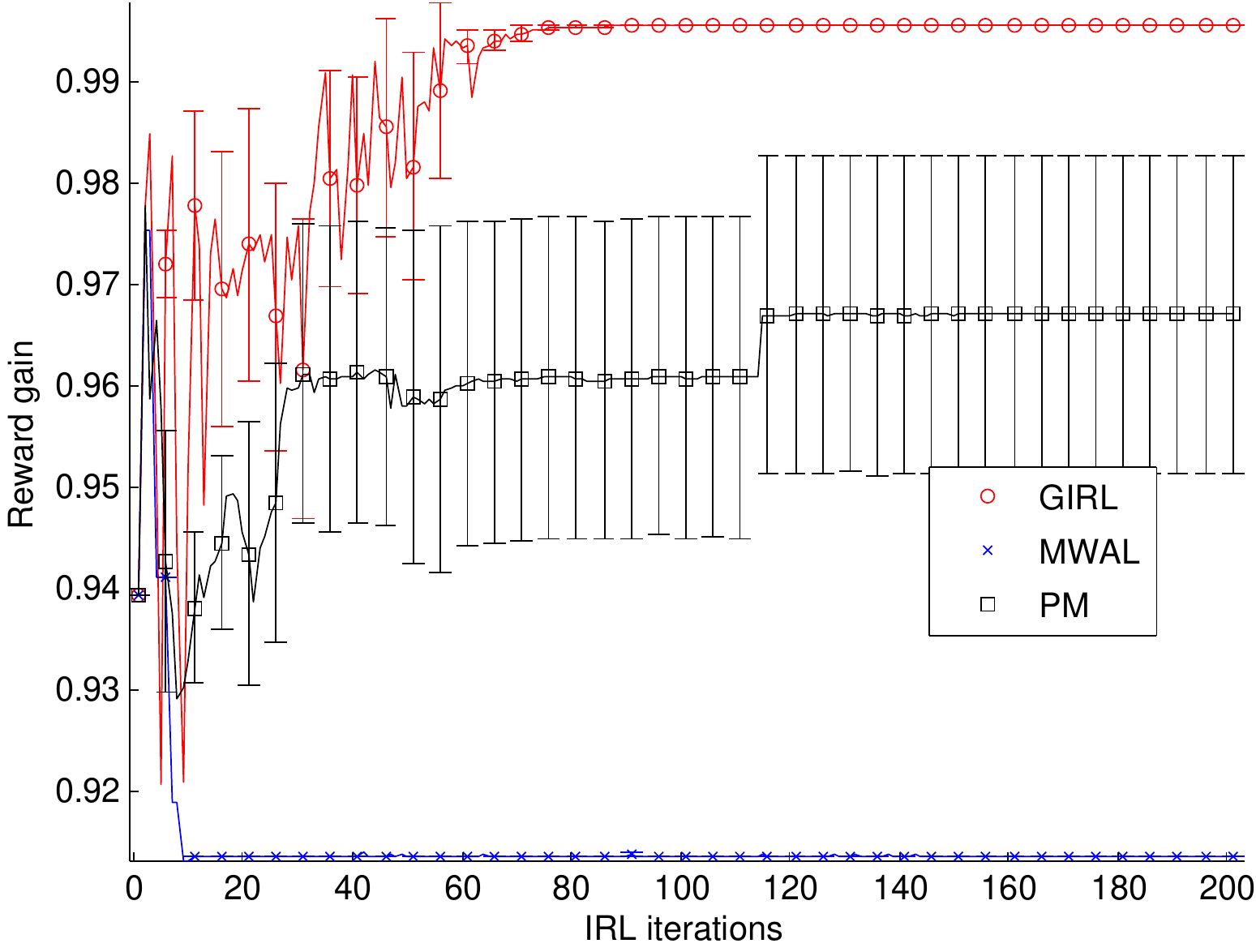}}

 \caption{Narrow passage. Average results over 10 runs. Top: policy similarity for (a) FP, (b) IA and (c) FP1.  Bottom: Accumulated reward for (d) FP, (e) IA and (f) FP1.}
\label{fig:narrowpassage}
\end{figure}

Figure \ref{fig:narrowpassage} show the results for the narrow passage problem with a macro-cell of size $2\times2$ states\footnote{Very similar results were obtained with bigger macro-cells which are not reported due to lack of space. For instance, for $4\times 4$ macro-cells, GIRL accumulated 0.98122.}. 
The results for the full derivative show that the three methods perform quite well obtaining an accumulated reward over.95 (compared to 0.99 of the expert's) (Fig. \ref{fig:narrowpassage}(d)), with GIRL achieving the best results (.987), almost identical to the expert. Similar results are obtained for the final estimated policy where GIRL behaves better than the other methods.  
The results almost do not change when using the approximate derivative (see Figs. \ref{fig:narrowpassage}(b) and (e)).  Again, GIRL obtains slightly better results than PM and MWAL.  In both cases, this is mainly due to some solutions getting stuck at local minima that make some states fall to the pits or fail to get out of the pit through the fastest route. Indeed, this is illustrated by the variance bars that are smaller for the GIRL method than for the others.   
For the FP1 approximation, GIRL behaves almost identically. PM degrades a little bit its performance and gets stuck in local minima more often. MWAL, on the other hand, is not able to recover any sensible policy. This is expected since that MWAL tries to match the feature expectations and, therefore, is more sensitive to approximations. However, it is surprising that it does work well with the IA approximation.

\begin{figure}[!t]
 \centering
\subfigure[Policy similarity, FP]{\includegraphics[width=0.3\textwidth]{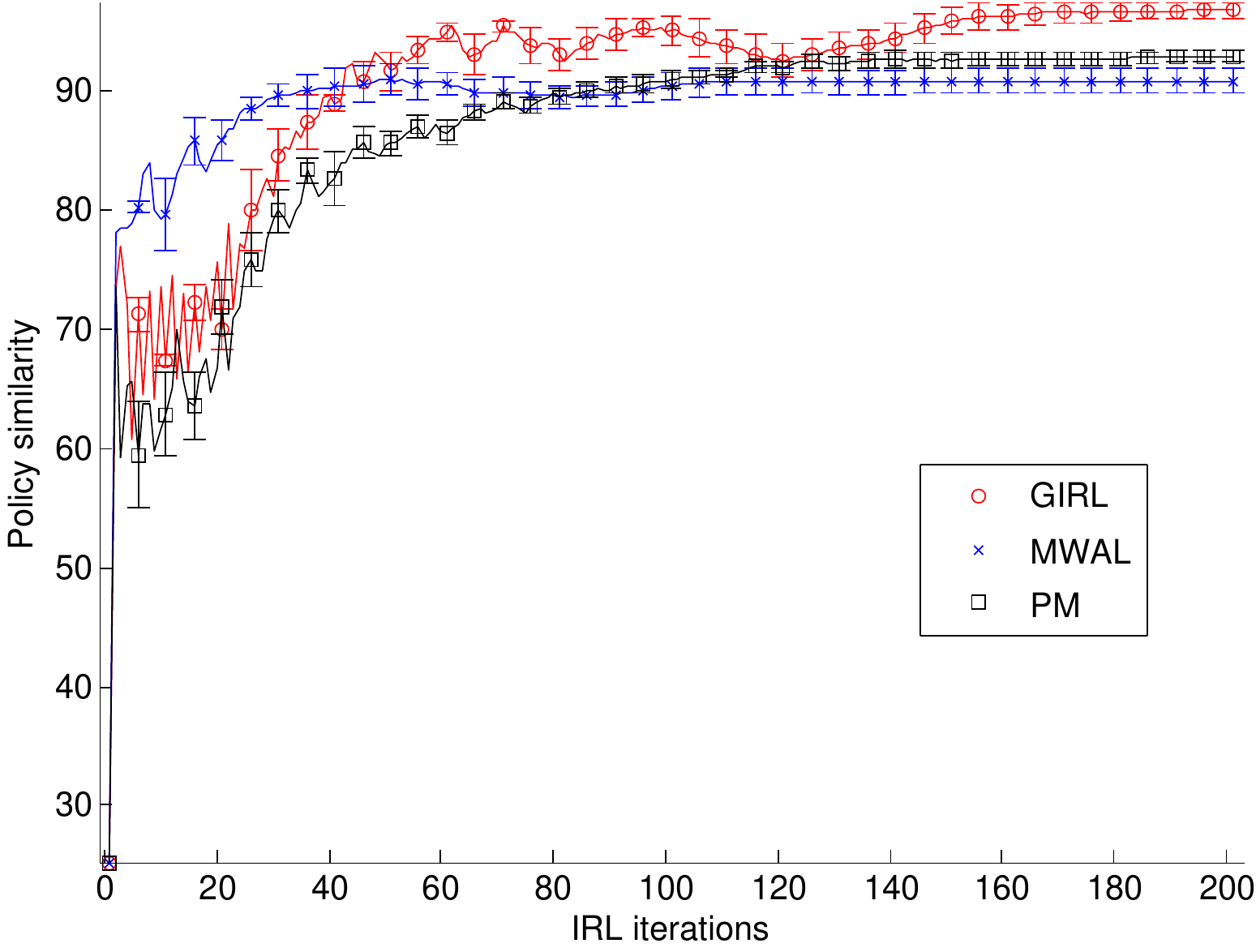}}
\subfigure[Policy similarity, IA]{\includegraphics[width=0.3\textwidth]{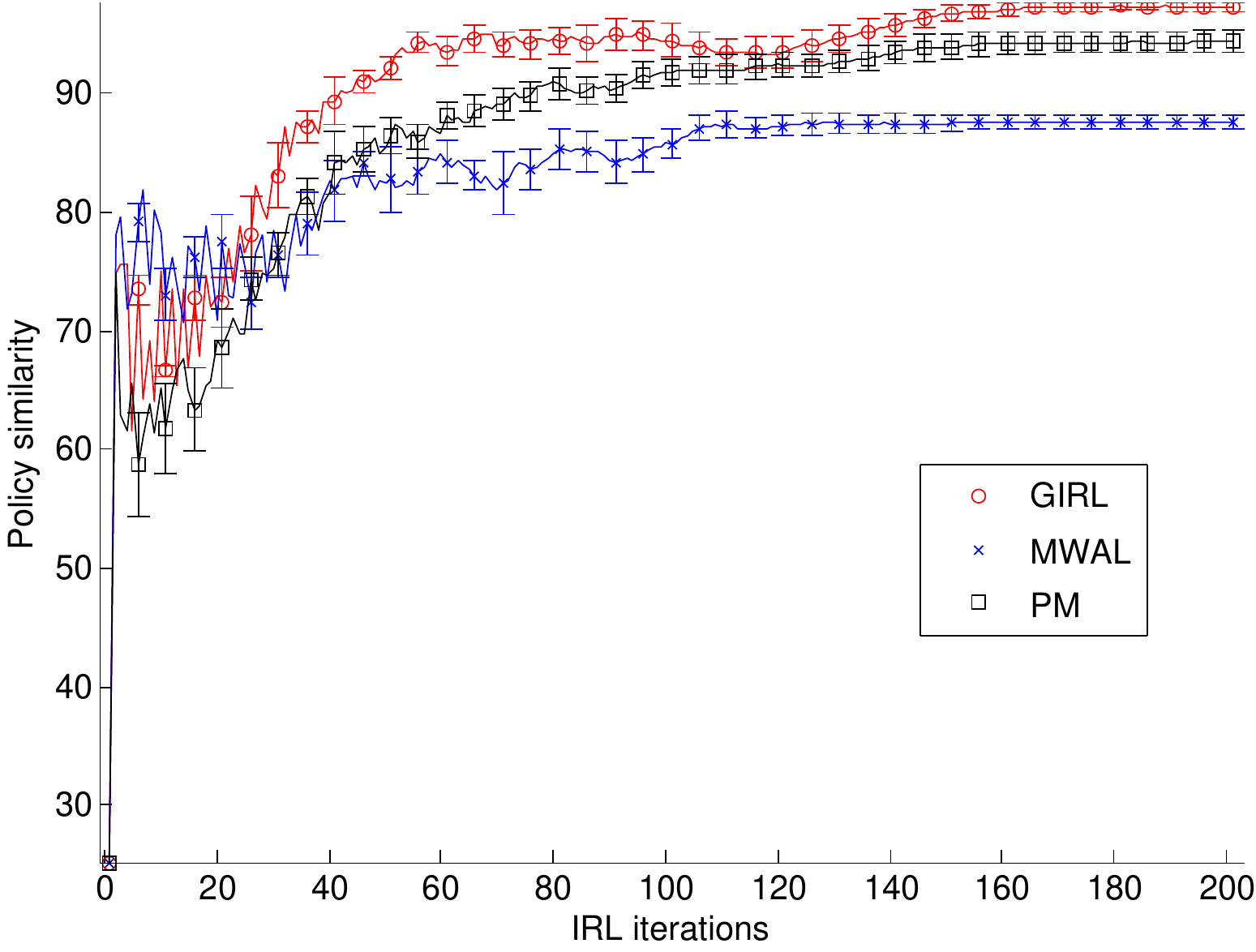}}
\subfigure[Policy similarity, FP1]{\includegraphics[width=0.3\textwidth]{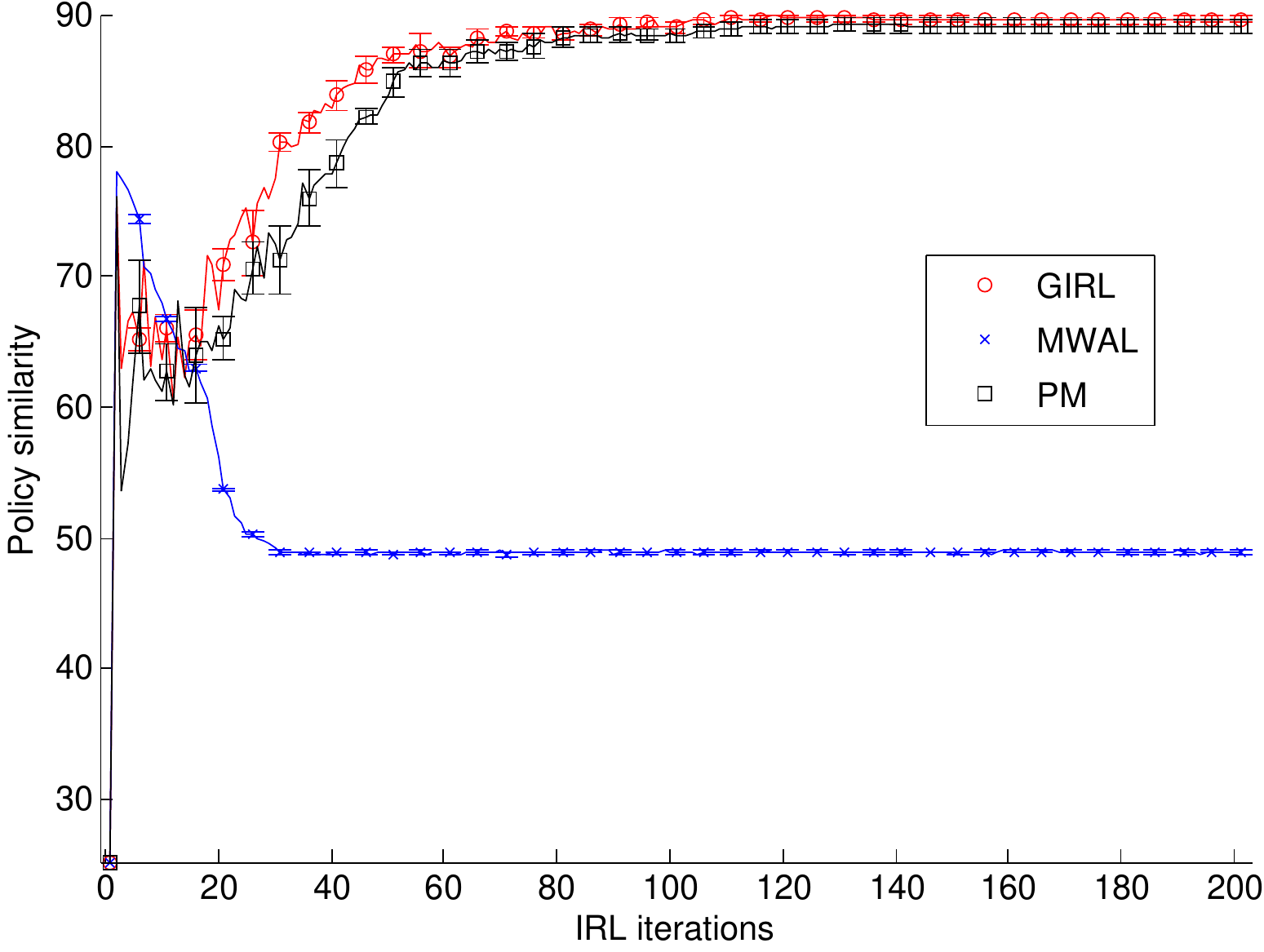}}

\subfigure[Acc. reward, FP]{\includegraphics[width=0.3\textwidth]{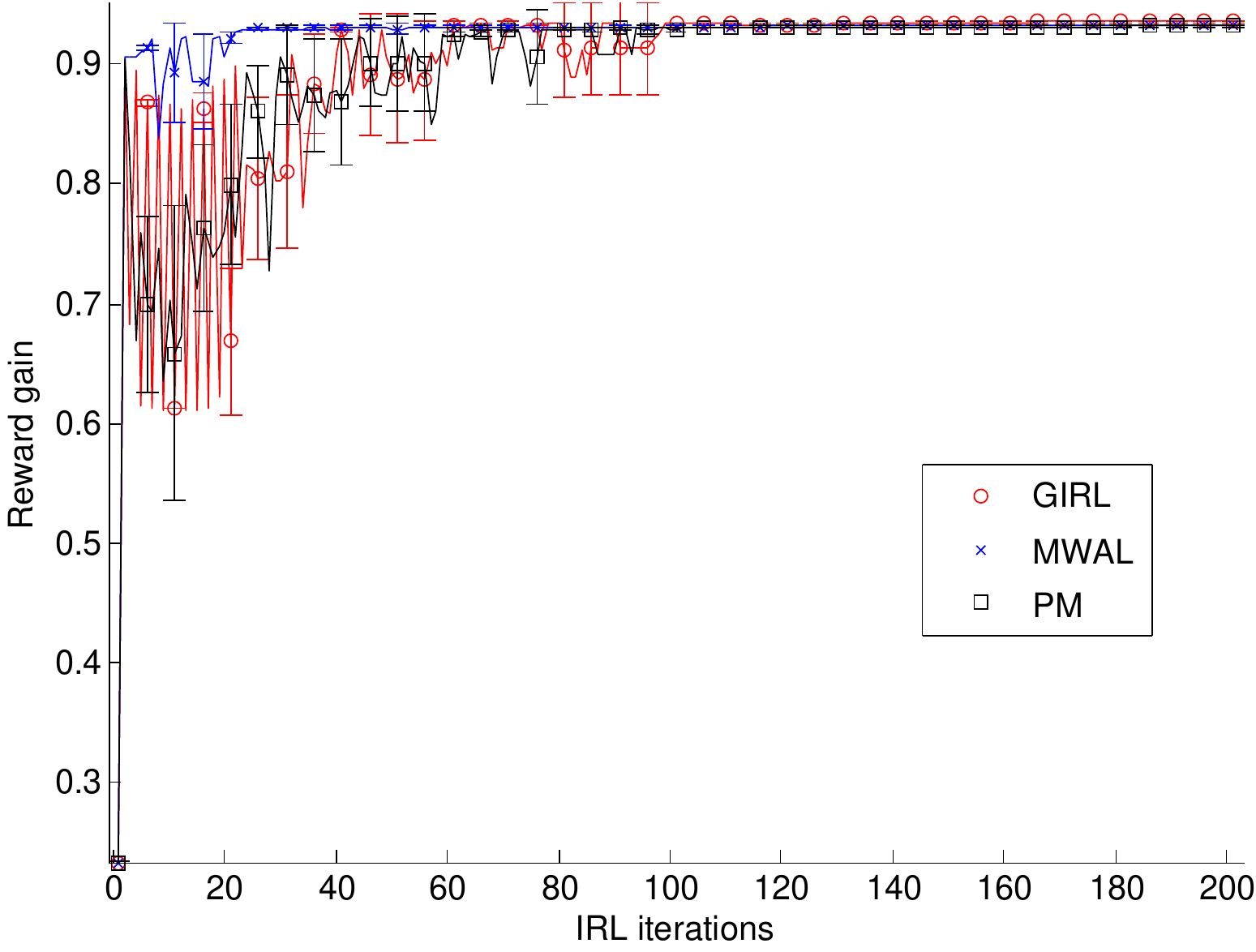}}
\subfigure[Acc. reward, IA]{\includegraphics[width=0.3\textwidth]{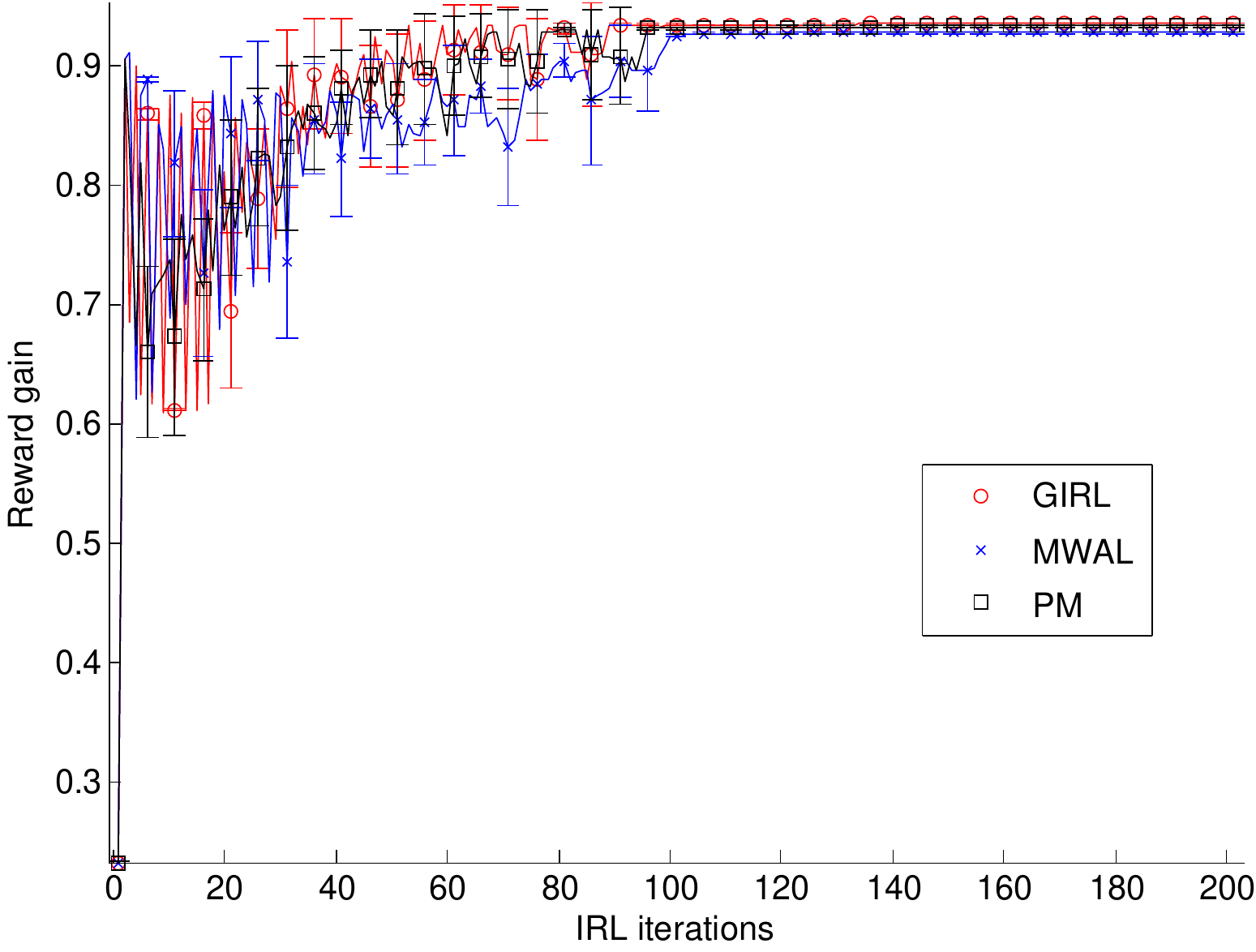}}
\subfigure[Acc. reward, FP1]{\includegraphics[width=0.3\textwidth]{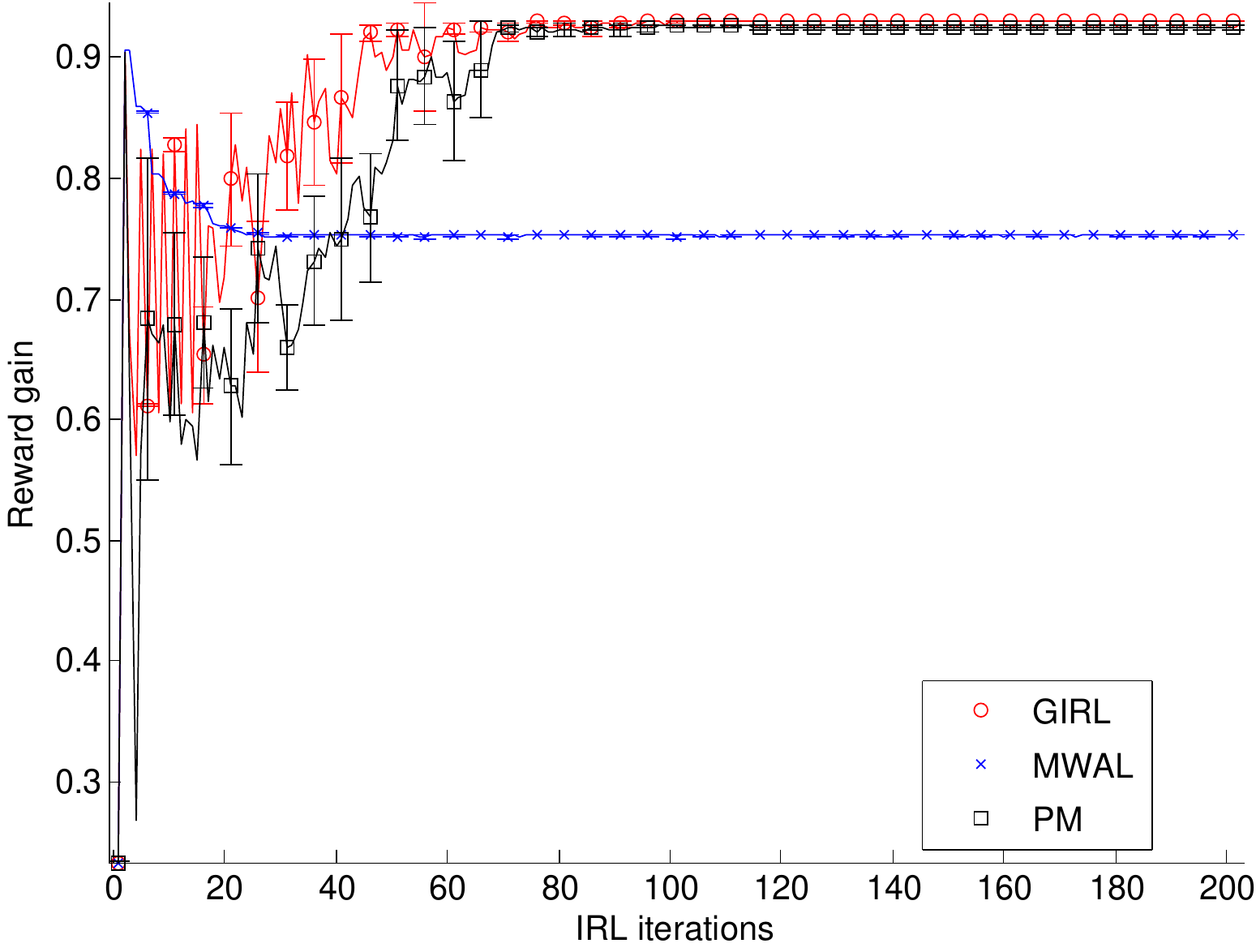}}
 \caption{Path problem. Average results over 10 runs. Top: policy similarity for (a) FP, (b) IA and (c) FP1.  Bottom: Accumulated reward for (d) FP, (e) IA and (f) FP1.}
\label{fig:paths}
\end{figure}

We now analyze the results for the path following problem (Fig. \ref{fig:paths}. Roughly, the results are the same as in the narrow passage. All the methods behave similarly for the FP and the IA cases with an accumulated reward almost identical to the expert (.93)\footnote{The accumulated reward with $4\times 4$ macro-cell was also best for GIRL (0.82). }. In fact, the differences between them are smaller than in the narrow passage. The main reason is that PM and MWAL do not seem to converge to worse local minima so often. This is indicated by a smaller variance in the figures, although it is still bigger than the GIRL's one. As in the previous case PM and GIRL almost keep the same performance with the FP1 approximation of the derivative and the MWAL algorithm fails to find a good solution. 

The difference in the computational cost varies enormously depending on the selected method to compute the derivative. For example, in the case of GIRL for the path following, the FP took 2.98 s per iteration on average, while the IA and the FP1 were 1.14 and 0.48 respectively. The results are consistent with the fact that FP has an exponential cost, while IA has a polynomial cost. On the other hand, the difference between IRL algorithms were negligible, being more important the number of iterations until convergence.


\subsection{Sailing}
In the problem of ``sailing'', proposed by Vanderbei \cite{Vanderbei}, the task is to navigate a sailboat from one starting point to a goal point in the shortest time possible. The speed of the sailboat depends on the relative direction between the wind and the sailboat, and the wind follows a stochastic process. In the MDP context, the state contains the data about the sailboat's position in the grid, the current wind direction, and the current tack. The possible actions are to move to any of the 8 adjacent cells in the grid. The rewards are negative, and corresponds to the time required to carry out the movement under current wind, i.e. it is faster to sail away from the wind than into the wind, and changing the tack induces an additional delay. The reward function is a linear combination of six features -away, down, cross, up, into, delay-, see \cite{Vanderbei} for details. These features depend on the state and the action. Thus, the problem is more challenging than the grid worlds which depend only on the state, that is, the location on the grid. The true weights we used in the experiments are the same ones used by Vanderbei \cite{Vanderbei}, namely $\theta^*=(-1,-2,-3,-4,-100000,-3)^T$.


\begin{table}
\centering
\begin{tabular}{|l|c|c|r|}
    \toprule
Method & $V_{R_E}^{\pi_{IRL}}$        & $\pi_E=\pi_\theta$   & T (s)      \\\midrule
 GIRL-FP &-11.76    & 93.87 & 332.44 \\
 MWAL-FP & -15440.31 & 83.87 & 522.59 \\
 PM-FP & -1630.48  & 90.90 & 376.26 \\
\hline  
GIRL-I & -11.76    & 93.91 & 52.73  \\
MWAL-I & -395.45   & 85.43 & 47.95  \\
PM-I & -2035.15  & 91.05 & 47.56  \\
\hline
GIRL-FP1 &-11.76    & 94.53 &34.53 \\
MWAL-FP1 & -16814.30 & 86.29 &35.06 \\
PM-FP1  & -416.44   & 93.91 &37.16 \\
        \bottomrule
    \end{tabular}\label{tabla:derivatives2}
   \caption{Experimental results for the sailing problem for 100 iterations of the GIRL, PM and MWAL methods using the fixed point method (FP), the independence assumption (IA) and a single step of the fixed point recursion (FP1). The reward accumulated by the expert is $V_{R_E}^{\pi_E}=-11.76$.} 
 	\label{tabla:derivatives}        
\end{table}

\begin{figure}[!t]
 \centering
\subfigure[Policy similarity, FP]{\includegraphics[width=0.3\textwidth]{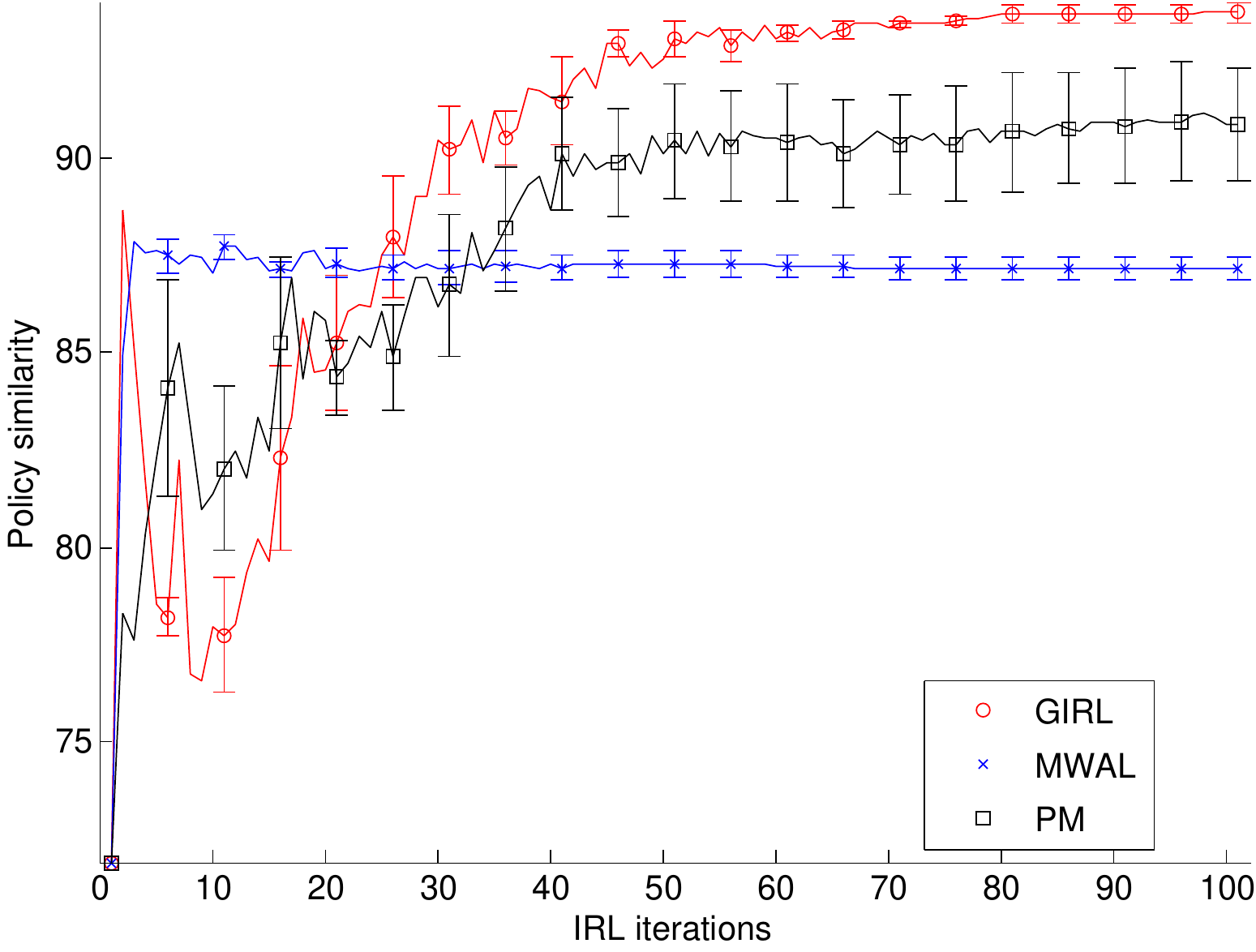}}
\subfigure[Policy similarity, IA]{\includegraphics[width=0.3\textwidth]{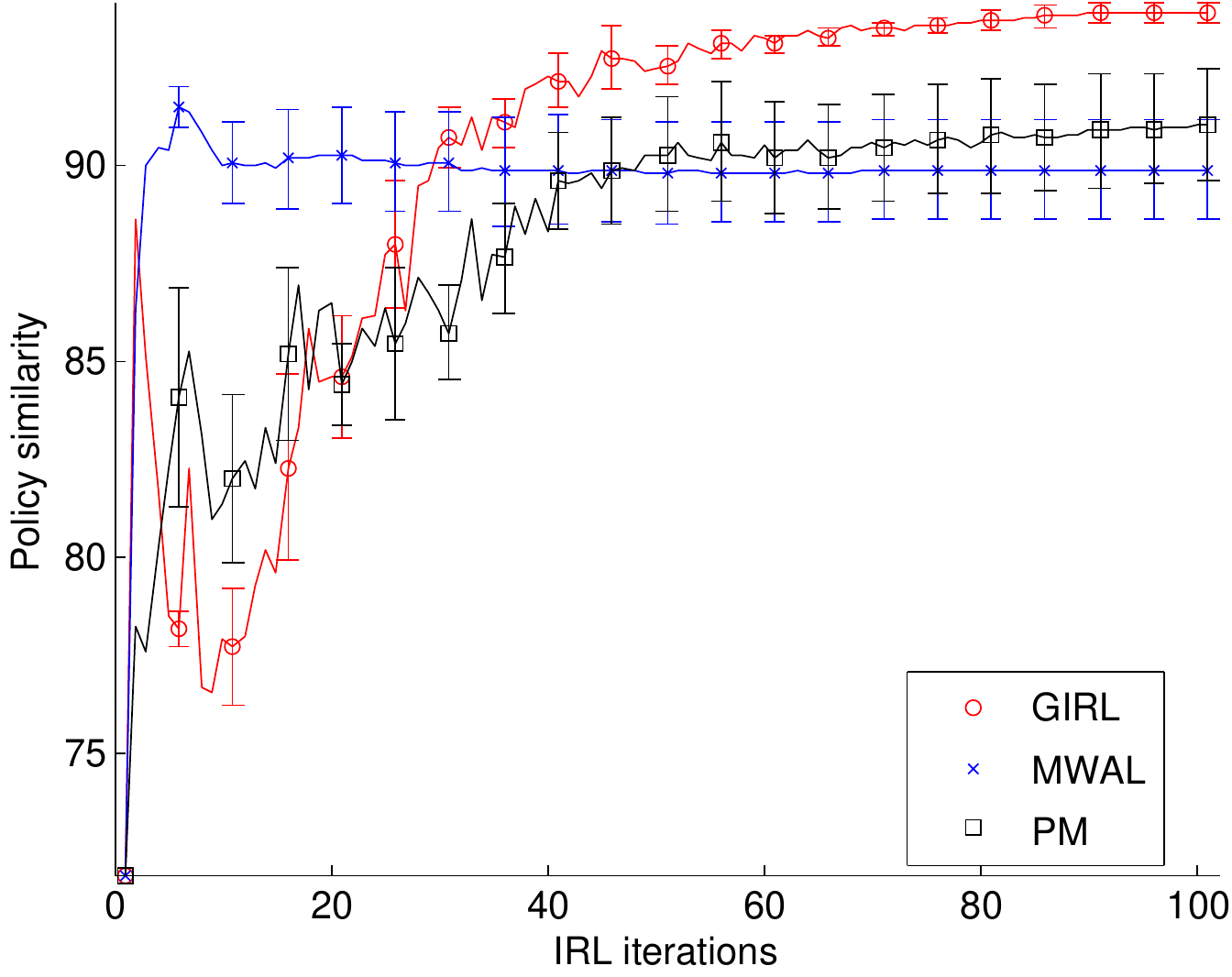}}
\subfigure[Policy similarity, FP1]{\includegraphics[width=0.3\textwidth]{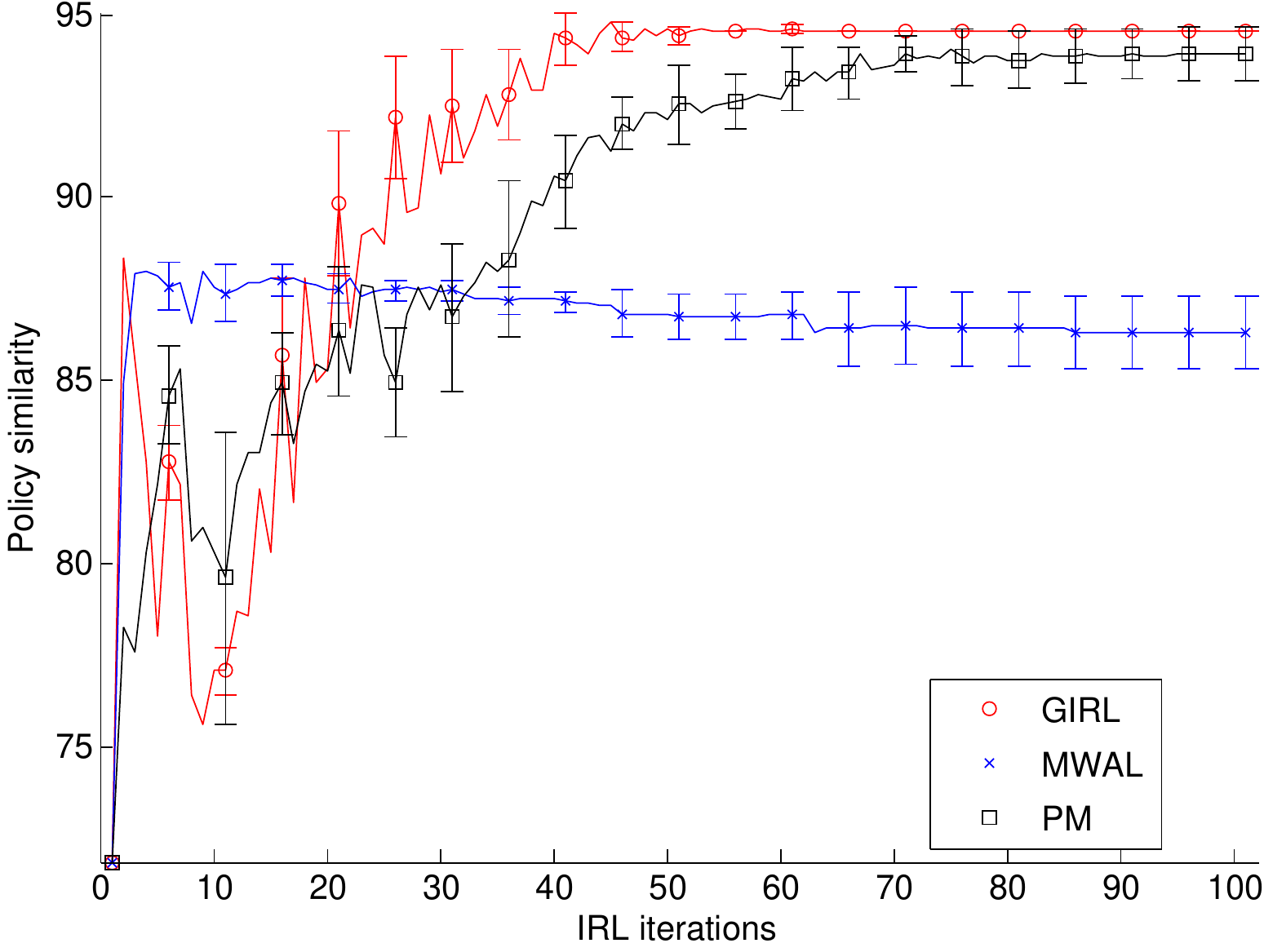}}

\subfigure[Acc. reward, FP]{\includegraphics[width=0.3\textwidth]{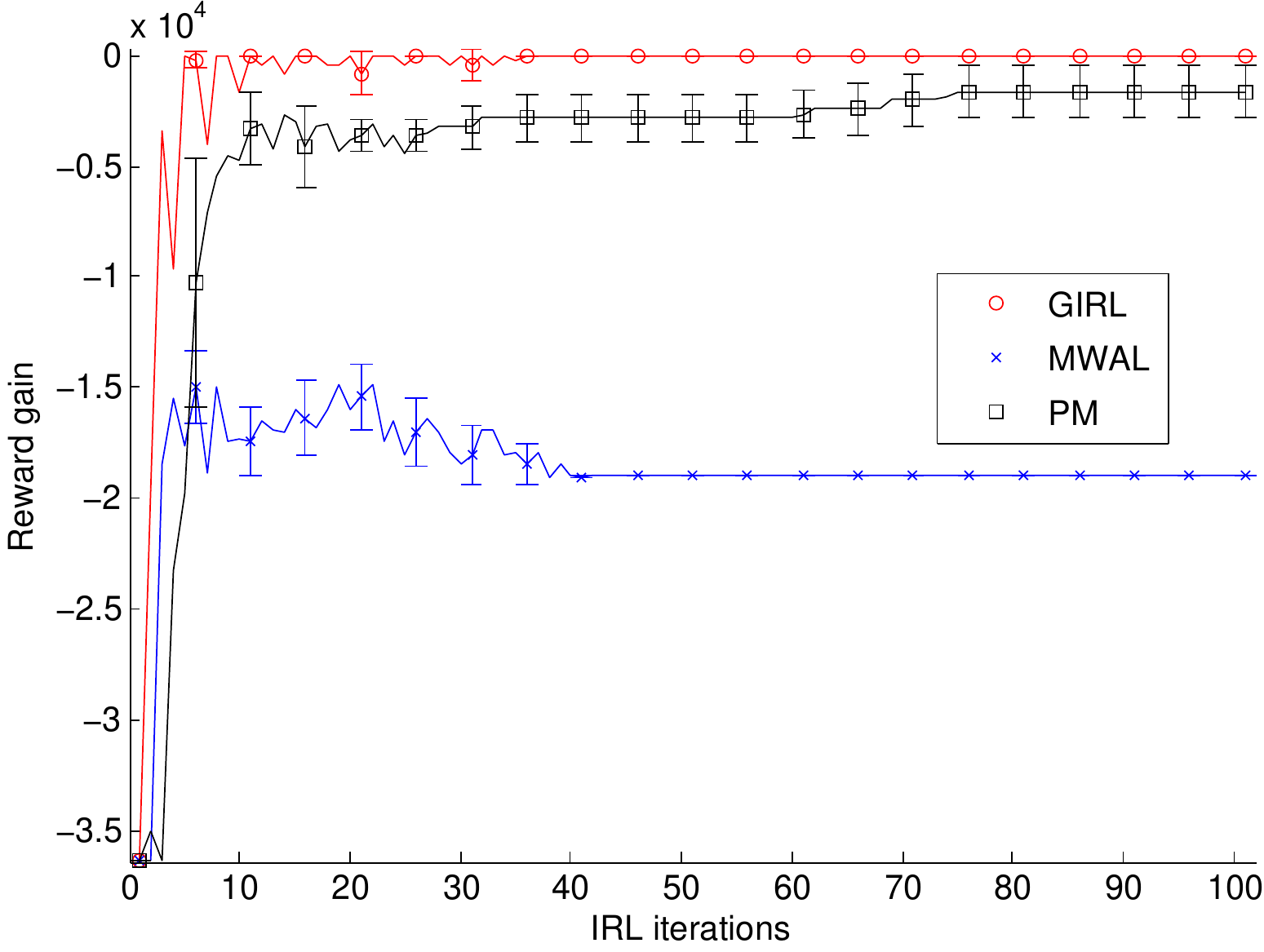}}
\subfigure[Acc. reward, IA]{\includegraphics[width=0.3\textwidth]{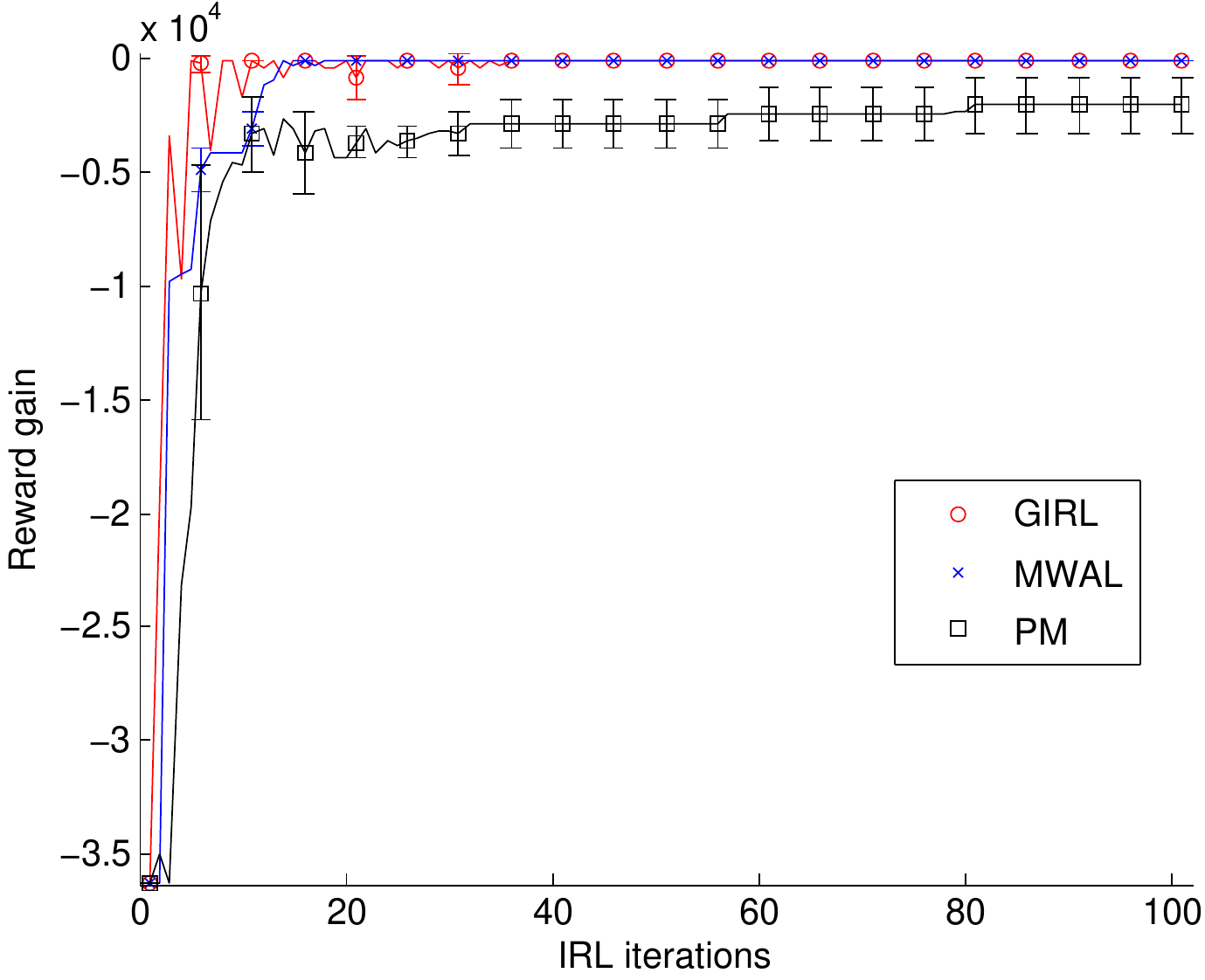}}
\subfigure[Acc. reward, FP1]{\includegraphics[width=0.3\textwidth]{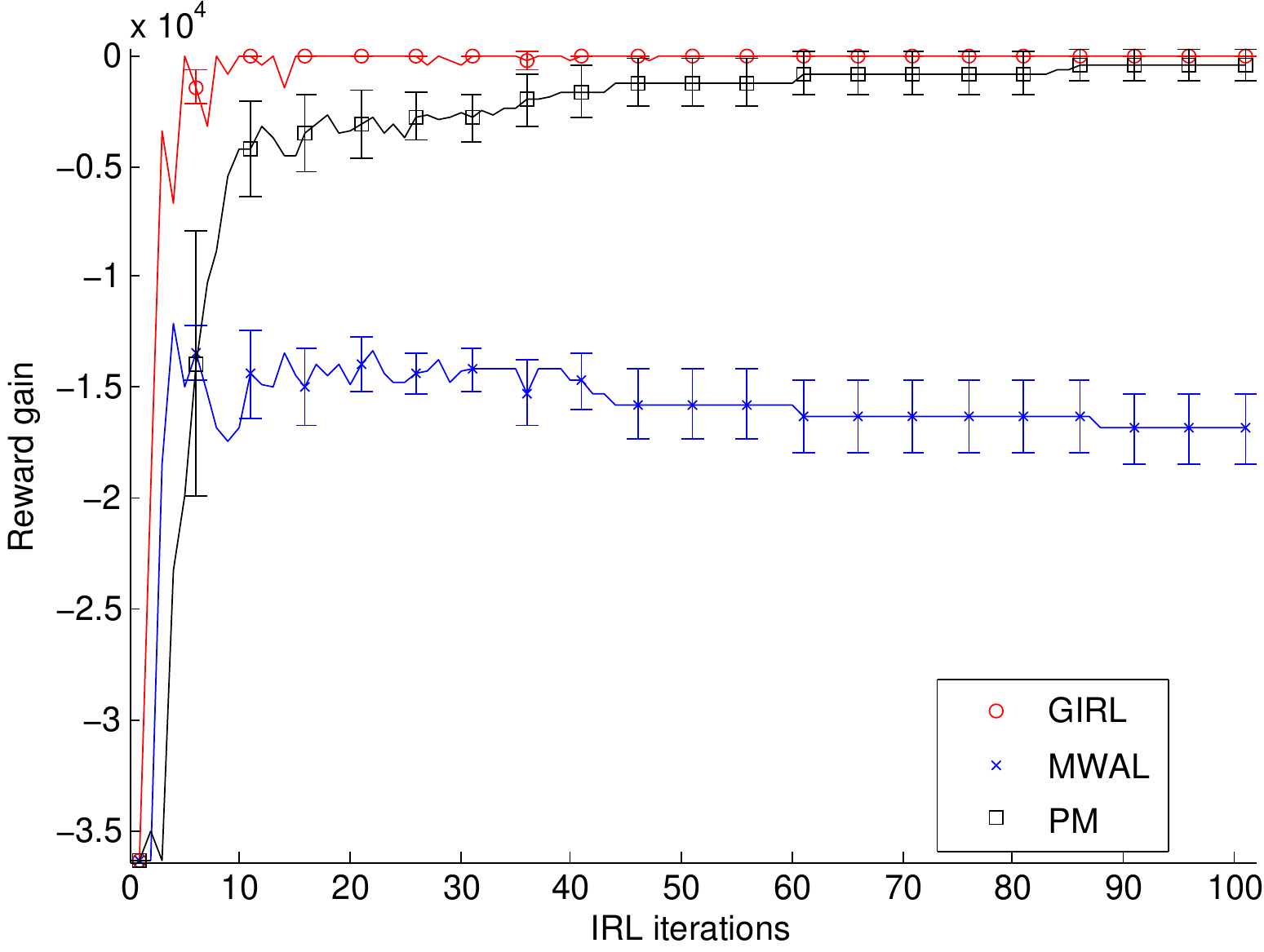}}

 \caption{Evolution of the performance of the different IRL methods in
the sailing problem: policy similarity and accumulated reward for (a,d) FP, (b,e) IA and (c,f) FP1.}
 \label{fig:plotsail}
\end{figure}

Table \ref{tabla:derivatives} summarizes the results of solving the sailing problem with different demonstrations averaged over 10 runs. The demonstrations consisted of 5120 expert trajectories each. We run the algorithms for one hundred iterations which was more than needed for convergence of the dissimilarity function $J(\pi_\theta, \mathcal{D})$ (around 40 IRL iterations). Figure \ref{fig:plotsail} shows the evolution of the two metrics, real accumulated reward of the learned policy $V_{R_E}^{\pi_{IRL}}$ and proportion of states $x_i$ where $\pi_E(x_i)=\pi_\theta(x_i)$. For this problem, the reward accumulated by the expert is $V_{R_E}^{\pi_E}=-11.76$. 

The GIRL and the PM algorithms perform reasonably well for the FP, IA and FP1 cases. GIRL had the best performance with an accumulated reward almost identical to the expert's one independently of the derivative used. PM have slight differences that could be due to sampling noise for the demonstrations. In general, it had a slightly smaller accumulated reward (between -2000 and -400) and also a higher variance than GIRL. Surprisingly, MWAL worked best in the IA case where it obtained an accumulated reward of -395. However, it did perform poorly in the other two cases where it did not converge to the right solution. This effect, which also appeared in the grid world, requires further investigation. 
In any case, the PM and MWAL solutions revealed that most of the loss in reward is due to a small number of wrong actions (going against the wind), which increased the final time. This was avoided by the GIRL method. 
Finally, the computational times depend mainly on the approximation used (see Table \ref{tabla:derivatives}) and, as in the grid world problems, they are very similar among methods.



\section{Conclusions}
\label{sec:conclusions}

In this paper we have reviewed gradient based algorithms for the problems of inverse reinforcement learning and apprenticeship learning. On one hand, we have discussed in detail the properties of a recently developed algorithm, the maximum likelihood IRL. We have shown that the probabilistic inference approach has connections with IRL as convex optimization. In particular, it represents an alternative cost function to the least squares criterion of the policy matching algorithm. The experimental results show that for some typical problems the behavior of the likelihood based algorithm is at least as good as the other methods. 
On the other hand, one of the most expensive steps in gradient based methods is the computation of the derivative of the policy. We have analyzed an approximation of the derivative that exploits the fact that \emph{small changes in the reward function do not affect the policy}. The approximated derivative can be computed, at every iteration, in polynomial time (instead of using a fixed point recursion). Results show that the obtained reward is much faster and as accurate as the ones obtained with the full derivative.


\bibliographystyle{plain}
\bibliography{irl}

\end{document}